\documentclass[fleqn,10pt]{wlscirep}
\usepackage[utf8]{inputenc}
\usepackage[T1]{fontenc}
\usepackage{amsmath,amsfonts}
\usepackage{algorithmic}
\usepackage{algorithm}
\usepackage{array}
\usepackage{textcomp}
\usepackage{stfloats}
\usepackage{url}
\usepackage{verbatim}
\usepackage{graphicx}
\usepackage{cite}
\usepackage{subfiles}
\usepackage{booktabs}
\usepackage{subcaption}  
\usepackage{multirow}
\usepackage{makecell}
\usepackage{siunitx}

\title{Therapist-Robot-Patient Physical Interaction \textit{is Worth a Thousand Words}: Enabling Intuitive Therapist Guidance via Remote Haptic Control}
\author[1,*]{Beatrice Luciani}
\author[1]{Alex van den Berg}
\author[1]{Matti Lang}
\author[1,2]{Alexandre L. Ratschat}
\author[1,2]{Laura Marchal-Crespo}

\affil[1]{Department of Cognitive Robotics, Delft University of Technology, Delft, The Netherlands}
\affil[2]{Department of Rehabilitation Medicine, Erasmus Medical Center, Rotterdam, The Netherlands.}

\affil[*]{bluciani@tudelft.nl}

\begin{abstract}
Robotic systems can enhance the amount and repeatability of physical guided motor training. Yet their real-world adoption is limited, partly due to non-intuitive trainer/therapist-trainee/patient interactions. To address this gap, we present a haptic teleoperation system for trainers to remotely guide and monitor the movements of a trainee wearing an arm exoskeleton. The trainer can physically interact with the exoskeleton through a commercial handheld haptic device via virtual contact points at the exoskeleton's elbow and wrist, allowing intuitive guidance. Thirty-two participants tested the system in a trainer–trainee paradigm, comparing our haptic demonstration system with conventional visual demonstration in guiding trainees in executing arm poses. Quantitative analyses showed that haptic demonstration significantly reduced movement completion time and improved smoothness, while speech analysis---using large language models for automated transcription and categorization of verbal commands---revealed fewer verbal instructions. The haptic demonstration did not result in higher reported mental and physical effort by trainers compared to the visual demonstration, while trainers reported greater competence and trainees lower physical demand. These findings support the feasibility of our proposed interface for effective remote human-robot physical interaction. Future work should assess its usability and efficacy for clinical populations in restoring clinicians’ sense of agency during robot-assisted therapy.
\end{abstract}
\begin{document}

\flushbottom
\maketitle
%
%
\thispagestyle{empty}
\section{INTRODUCTION}

Robotics is paving the road towards high-intensity~\cite{Mehrholz08}, personalized~\cite{MarchalCrespo2009}, and objectively quantifiable rehabilitation of neurological patients~\cite{Longatelli2023}, bypassing current limitations in therapists' availability and endurance~\cite{Richard2020}. Yet, despite years of technological advancements in the development of rehabilitation devices, therapists often express concerns regarding the technology's usability, intuitiveness, and the loss of direct physical contact with the patient~\cite{Lo2020, LucianiTAM}. In conventional rehabilitation, therapists rely on a variety of communication channels, such as physical (haptic), visual, and/or verbal communication, to guide and correct patients’ movements~\cite{Platz2021}. Many robotic devices, however, mediate this therapist-patient sensory interaction, reducing the therapists' ability to intervene dynamically during training, limiting their role to mere observers~\cite{Hasson2023}. Even in the few solutions that enable therapists to interact with and act on the robotic device, this is done by allowing the tuning of high-level robotic parameters---think of percentage of arm-weight compensation or movement speed~\cite{Molle2025, Stephenson2017}---, and generally lack the intuitiveness of conventional therapy approaches~\cite{Celian2021}. This superficial interaction diminishes therapists' sense of control and involvement in the rehabilitation session~\cite{LucianiTAM}, leading to a negative perception of the robot as a constraint rather than a partner, which in turn hinders the widespread acceptance of robotic systems in clinical practice~\cite{LucianiTAM}. 

To enhance the acceptability of rehabilitation technology, research has focused on developing therapist-in-the-loop interfaces, which aim to restore therapists' sense of agency by facilitating a better understanding of the robot behavior~\cite{Sommerhalder2022} or providing new opportunities for interaction with the robot~\cite{Pavón-Pulido2020}. Recent studies have shown that the robot behavior---e.g., robot joint trajectories, physical support profiles---can be effectively adjusted by the physical intervention of external users~\cite{Amato2024}, such as a therapist, improving the intuitiveness and therefore acceptability~\cite{Luciani2023} of robotic rehabilitation. Building on the new opportunities given by physical therapist-robot interactions, recent work has emphasized the need to integrate therapists within the control loop through robot-mediated physical human-human interactions, where the therapist and patient can interact, physically, via robotic devices~\cite{Vianello2025}. The idea is to give therapists chances for more direct and natural (i.e., more ``low-level") control over the robotic device during the therapy session, preserving their clinical intuition~\cite{Vianello2025, Hasson2023}.

Few teleoperation systems have been proposed that allow therapists to intervene intuitively---rather than be sidelined---mimicking their classic approach to rehabilitation even with the presence of a robot. For instance, Hou \textit{et al.}~\cite{hou2025} introduced a tele-rehabilitation framework that uses two end-effector devices (non-rehabilitation intended 7-DoF Franka Emika Panda) to transfer force inputs from a participant acting as a therapist (denoted here as trainer) to the hand of an unimpaired participant acting as a patient (trainee) and adjust their trajectories. This system enables the trainer to correct trainees' movements without requiring the trainer to reprogram the reference trajectory or adjust abstract parameters (from the trainer's point of view). Its control system, although effective in replicating trainer-informed adjustments, acts only at the robot end-effector level, i.e., the hands of the trainer and trainee. This limits the potential therapist's chances for interaction with patients' individual joints and could potentially produce incorrect/unhealthy arm poses. This limitation is overcome in the ``Beam-Me-In'' approach from Baur and colleagues~\cite{baur2019beam}, where both the trainer and the trainee used ARMin exoskeletons located in the same room. The system was used to transmit three-dimensional (3D) forces from the ``patient''---unimpaired participants performing prerecorded impaired arm movements---to the therapist, allowing the clinicians to physically perceive the patient's movement limitations, e.g., range of motion limits, spasticity. While clinicians appreciated the system as an effective assessment tool, it is important to highlight that it was not designed to provide active correction or physical guidance to patients during training. Additionally, the setup requires two complex robotic systems, which can be excessively expensive for a clinic. Importantly, the therapists should learn how to feel the forces reflected from the robot worn by the patient while wearing an exoskeleton themselves, stepping out from their usual point of view, i.e., perceiving forces through their hands while interacting with patients at different contact points. In summary, while preliminary research provides a good foundation for further development, it falls short in either not providing enough support for patients' individual joints or being too complex from the therapist's perspective, who cannot act upon the movements of their patients. To our knowledge, no study has combined all the following elements in a single teleoperation system: (i) a simple, intuitive end-effector haptic interface for trainers, (ii) the ability to deliver corrective or assistive inputs to exoskeleton-supported trainees, and (iii) the projection of the trainee–robot interaction dynamics back to the trainer for real-time assessment.

In this study, we address this gap by introducing a therapist-in-the-loop teleoperation system that utilizes a haptic end-effector device with reaching and grasping capabilities (Sigma.7, Force Dimension, Switzerland) that therapists can use to haptically feel patients' movements and ``physically'' interact with an exoskeleton (ARMin, SMS lab, ETH Zurich) to physically adjust their arm poses. To facilitate the therapists' visualization of the exoskeleton they are remotely interacting with, we created a virtual replica of the exoskeleton (digital twin~\cite{ratschat2023}) and a generic avatar of the patient, which the therapist can visualize from a first-person perspective in immersive virtual reality (IVR) using a commercial Head Mounted Display (HMD). We decided to use an IVR over a screen to facilitate the visualization of the required 3D movements to guide the exoskeleton pose~\cite{wenk, wenk2023effect}. Through grabbable contact points---at the virtual exoskeleton elbow and wrist joints---therapists can haptically feel and adjust the patient’s arm configuration across multiple joints, overcoming the constraints of end-effector-only interaction and enabling intuitive and precise physical assistance. 

To evaluate the feasibility of our teleoperation training haptic system, we performed a within-subject experiment with 36 unimpaired young participants, tested in pairs. Each dyad consisted of a \textit{Trainer} (therapist role) and a \textit{Trainee} (patient role); the Trainer guided the Trainee’s exoskeleton-controlled arm toward target arm poses using the proposed haptic system. 
The proposed approach was compared, both in terms of performance and subjective experience, to a conventional remote visual demonstration of the target pose (using a webcam at the trainer's side and a screen on the trainee side), reflecting the standard method used in conventional therapy when therapists show patients a pose to be replicated~\cite{Christensen2023}. Dyads were allowed to orally communicate freely. We expected that the proposed therapist-in-the-loop system would enable trainers to instruct trainees in reaching the desired arm configuration with faster and smoother movements than the conventional remote visual demonstration, while reducing the need for complementary (i.e., verbal) interactions. We also expected the teleoperation haptic system to increase the effort required from the trainer, but, in turn, enhance their sense of competence, both of which were assessed through questionnaires.
\section{METHODS}

\subsection{Experimental System: Teleoperation Therapist-in-the-loop Framework}
\subsubsection{System Architecture}

\begin{figure*}[h!]
    \centering
    \includegraphics[width=0.8\linewidth]{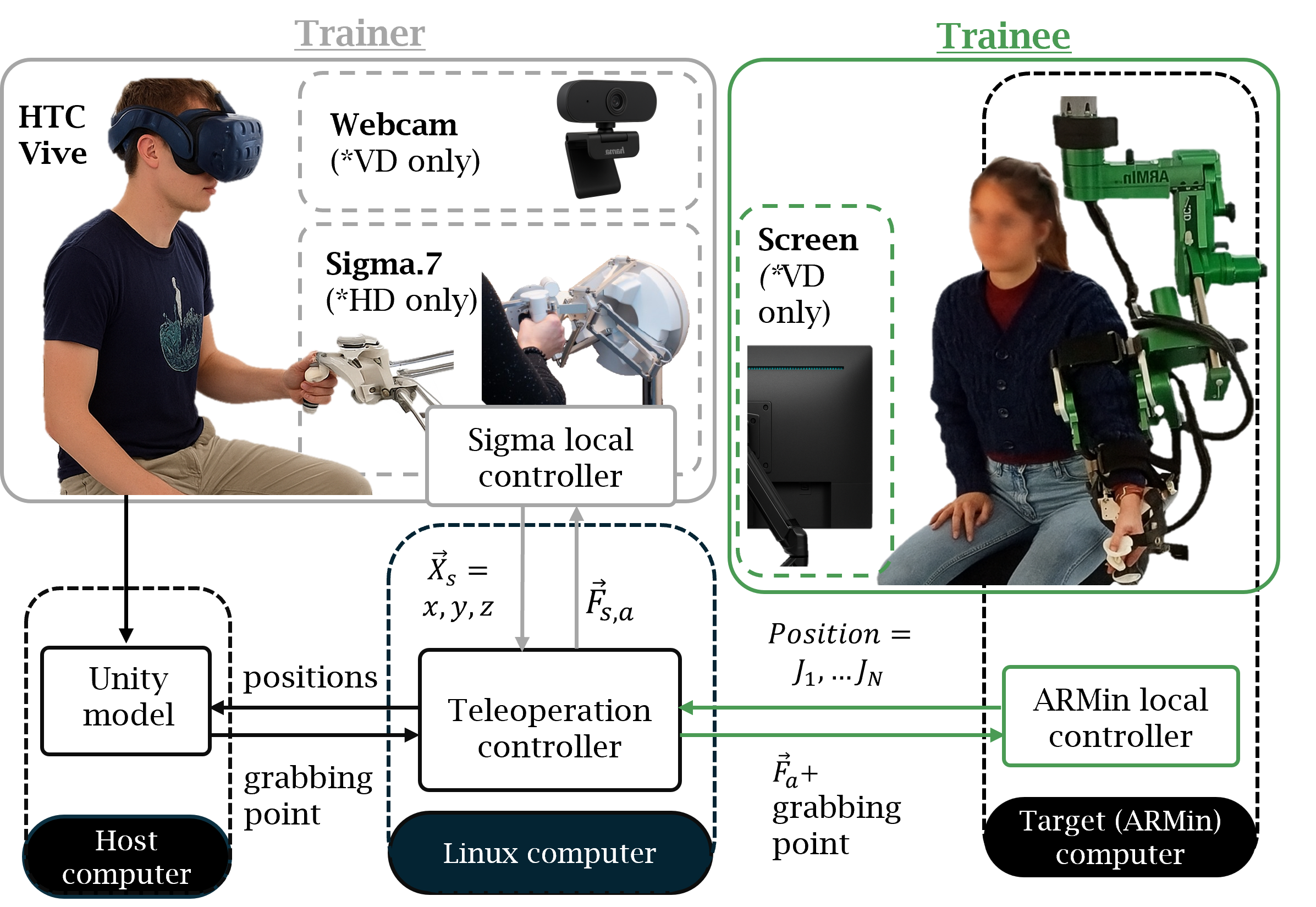}
    \caption{Overview of our teleoperation therapist-in-the-loop haptic system. The Trainee (right) wears the ARMin exoskeleton, which is remotely controlled by the Trainer (left) using a Sigma.7 haptic device. The Trainer wears a headset and is immersed in VR, where the digital twin of the exoskeleton and the Trainee’s avatar are visualized. The proposed haptic framework is evaluated in the Haptic Demonstration (HD) condition and compared with a conventional Visual Demonstration (VD) approach. In the VD condition, the Trainer is provided with a webcam, and the Trainee visualizes the Trainer’s movements on a screen in real time. The system includes three interconnected computers, represented in the figure through three rounded rectangular blocks labeled “Host computer”, “Linux computer”, and “Target (ARMin) computer”. In the image, $\vec{X_s}$ denotes the Cartesian position of the Sigma.7 end-effector, $\vec{F_{s,a}}$ is the interaction force computed on the Sigma.7 side by the Teleoperation Controller, $\vec{F_{a}}$ the force computed on the ARMin side, and $J_1,\dots, J_N$ are the ARMin joint positions recorded in real-time.}
    \label{fig:setting}
\end{figure*}

Our therapist-in-the-loop system (Fig.~\ref{fig:setting}) comprises two robotic devices: one for the Trainer and one for the Trainee, which communicate through a teleoperation controller. 
The Trainer's system side consists of:
\begin {itemize}  
    \item The Sigma.7 (Force Dimension, Switzerland), a haptic end-effector device. It provides 7 Degrees of Freedom (DoFs), namely three translational (moving in space), three rotational (orienting the hand), and one for grasping~\cite{forcedimension}. The Sigma.7 can provide forces to the hand of up to, respectively, \SI{20}{\newton}  in translation and \SI{8}{\newton} in grasping, within a workspace of $\varnothing$ \qtyproduct[product-units = single]{190 x 130}{mm}.
    \item The HTC Vive Pro Eye Head Mounted Display (HTC Vive, Taiwan \& Valve, USA).
\end{itemize}
The Trainee's system side consists of:
\begin{itemize}
    \item The ARMin~\cite{just2018exoskeleton} upper-extremity exoskeleton, with six active DoFs, four at the shoulder and elbow, and two at the forearm. All DoF are equipped with encoders (Maxon Encoder MR 1000, Maxon Group, Switzerland) and redundant analog sensors to accurately measure 
    the exoskeleton joint angles. The human-robot interaction forces/torques were measured using three force/torque sensors (Mini45, ATI Industrial Automation, Apex, USA) located beneath the upper and lower arm cuffs and hand module.
\end {itemize}

The conventional remote visual demonstration (VD) system included a webcam (Hama, 2Views, Denmark) at the Trainer's side and an LED screen (\SI{109}{\centi\metre}, 43UD79, LG, South Korea) located in front of the Trainee.

%

The system includes three interconnected computers (as indicated in Fig.\ref{fig:setting}): (1) \textit{Host} computer (OS: Windows 10, RAM: 32 GB, NVIDIA GeForce RTX 4070 Ti), (2) \textit{XPC-Target} computer (Spectra GmbH \& Co. KG, RAM: \SI{2}{\giga\byte}, CPU: Intel Core 2, \SI{3}{\giga\hertz}), and (3) \textit{Linux computer} (OS: Ubuntu 22.04, RAM: \SI{32}{\giga\byte}, NVIDIA GeForce RTX 2080 Ti, CPU: AMD Ryzen Threadripper 2950x 16-core processor, \SI{3.5}{\giga\hertz}).

The ARMin exoskeleton is controlled by the \textit{XPC-Target} computer, which runs a Simulink model started from the \textit{Host} computer. The \textit{Host} computer receives ARMin state data from the \textit{Target} computer (through the \textit{Linux} computer), namely each joint angular position $J_1,\dots,J_6$, which are employed to animate the ARMin digital twin in Unity (see Sec.~\ref{sec: virtual env}). The Trainer can interact with this digital twin to manipulate the position of the real exoskeleton through the Sigma.7 controlled through the separate \textit{Linux} computer. The latest manages the teleoperation control (see Sec.~\ref{sec: teleoperation}) and is connected to the Sigma.7 interface via USB cable. Communication from the \textit{Linux} machine to both the \textit{Host} and the \textit{Target} computers is done using a custom, single-threaded C++ User Datagram Protocol (UDP) at approximately \SI{450}{\hertz}.

\subsubsection{Immersive Virtual Environment}
\label{sec: virtual env}
\begin{figure}[h]
    \centering
    \includegraphics[width=\textwidth]{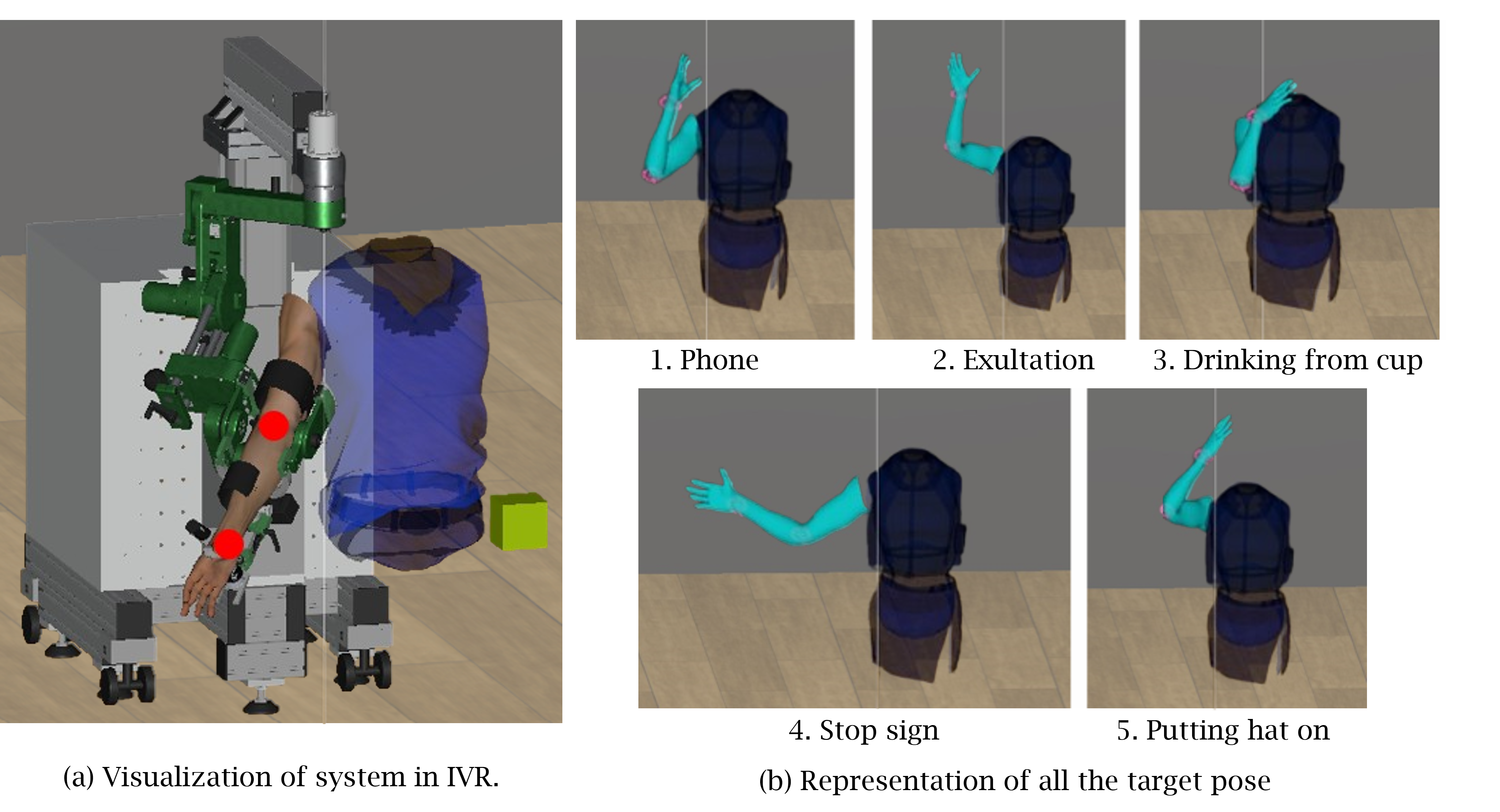}
    \caption{(a) Remote visualization of the virtual environment displayed to the Trainer. The visualization includes the ARMin digital twin~\cite{ratschat2023} and a representation of the Trainee's arm and torso (male or female). The red spheres on the avatar's wrist and elbow joints represent the graspable points, and the yellow cube is the representation of the end-effector of the Sigma.7 manipulated by the Trainer. (b) Snapshot of the five target arm poses that the Trainee was guided to execute during the experiment.}
    \label{fig:system_overview}
\end{figure}

The virtual environment the Trainer visualizes through the HMD was developed in Unity (version 2022.2.15f1, Unity Technologies, USA). It includes the ARMin digital twin developed and validated by Ratschat \textit{et al.}~\cite{ratschat2023} (Fig.~\ref{fig:system_overview}a). We also included a representation of the Trainee's right arm and torso (either in a male or female version to match the trainee's preferences) using the \textit{First Person Generic Arms Pack} and \textit{Adventure Character} packages from the Unity asset store. The \textit{Final IK} package from RootMotion was used to match the movement of the Trainee's virtual arm with the ARMin digital twin. 

The visualization of the Sigma.7 in the virtual environment was simplified to a unique yellow cube (Fig.~\ref{fig:system_overview}a), which translates along the $x$, $y$, and $z$ directions following the movement of the Sigma.7 end-effector (1:1 mapping).  
Through the Sigma.7 gripper, the Trainer can virtually ``grab'' the ARMin digital twin at predefined points depicted in the virtual environment with two red spheres (Fig.~\ref{fig:system_overview}a). Inspired by the way therapists interact with their patients, we limited the predefined points to two: the wrist and elbow joints. Grabs can only be executed when the virtual representation of the Sigma.7 end-effector, i.e., the yellow cube, is within a range of \SI{10}{\centi\meter} from any of the predefined points, i.e., red spheres. To inform the Trainer that a ``grab'' was successful, the red sphere turns green and only returns to red when the Trainer releases the gripper.

\subsubsection{Teleoperation Controller}
\label{sec: teleoperation}
The teleoperation controller computes the forces generated when the Trainer interacts with the Trainee at the graspable points in the virtual environment as a means to project those forces back to the Trainer (through the Sigma.7) and the Trainee (through ARMin). The two devices were positioned and oriented at different locations within the virtual environment, and therefore, the first step consisted of representing the Sigma.7 end-effector pose $\vec{X_{s}}$ obtained from the Force Dimension SDK into the ARMin coordinate system, following the equation:. 
\begin{equation}
\vec{X'_{s,a}} = \vec{X_{s}} \cdot \mathbf{R}_{S,A} \cdot C_{scale} + \vec{X_{o}} ,
\label{eq:transformation}
\end{equation}
where $\mathbf{R}_{S,A}$ is a rotational matrix around the vertical z-axis of $\theta = \text{\qty{45}{\degree}}$ and $\vec{X_{o}} = [0.34, 0, 1] m$ the translational offset between frames determined experimentally to ensure a precise overlay of the two devices' workspaces. 
Moreover, the scaling factor ($C_{scale}$ = 10) was added to account for the smaller workspace of the Sigma.7 compared to that of ARMin. 

The forces applied to the Sigma.7 ($\vec{F}_{s,a}$) and the ARMin ($\vec{F}_{a}$) as a consequence of virtual interactions were calculated based on the difference in position and velocity between the Sigma.7 end-effector and the interacted graspable point (i.e., exoskeleton wrist or elbow joint) of the ARMin ($\vec{X_{a}}$):
\begin{equation}
    \vec{F}_{s,a} = -P_{s} (\vec{X'_{s,a}} - \vec{X_{a}}) - B_{s} (\dot{\vec{X'_{s,a}}} -\dot{\vec{X_{a}}}) ,
    \label{equ:force_sigma}
\end{equation}
\begin{equation}
    \vec{F}_{a}  = -P_{a} (\vec{X_{a}} - \vec{X'_{s,a}}) - B_{a} (\dot{\vec{X_{a}}} -\dot{\vec{X'_{s,a}}}) .
    \label{equ:force_armin}
\end{equation}

The proportional and derivative gains were fine tuned by trial and error to $P_{s} =$ \SI{30}{\newton\per\metre} and $B_{s}=$ \SI{0.1}{\newton\second\per\metre} for the Sigma.7 and to $P_{a} =$ \SI{80}{\newton\per\metre}  and $B_{a} =$\SI{4}{\newton\second\per\metre}  for the ARMin.

The calculated forces were then transformed back to the respective devices. In particular, the forces obtained for the Trainer side were rotated back to the Sigma.7 reference frame by $\theta$. 
The interaction forces determined for the ARMin, either at the wrist or elbow joint contact points, were transformed into joint torques using the corresponding Jacobian of the exoskeleton. The torques were then added into the device baseline controller, i.e., a zero-torque controller from Dalla Gasperina \textit{et al.}  \cite{dalla2023quantitative} that allows a transparent interaction with the exoskeleton when no corrective forces from the Trainer are applied.

\subsection{Feasibility Test: Comparing Haptic and Visual Movement Demonstration}

\subsubsection{Participants}
To evaluate the feasibility of the proposed teleoperation therapist-on-the-loop haptic framework, we conducted an experiment with thirty-two healthy young participants (17 male, 15 female; mean age, 26 years; age range, 23-33 years). 
All participants, mostly students enrolled in university or higher professional education programs, provided written consent to participate in the study.
The study was approved by the Human Research Ethics Committee of Delft University of Technology (HREC, Application ID 3928).

\subsubsection{Experimental Task and Protocol}

We compared the effectiveness of the Trainer in communicating upper-limb desired poses to the Trainee using our haptic framework against a gold standard real-time video streaming approach. 
The experimental task consisted of guiding the Trainee towards a set of five predefined arm poses, resembling common activities of daily living (ADL)~\cite{schwarz2022characterization}: exultation, drinking from a cup, answering the phone, putting a hat on, and a stop gesture (Fig. \ref{fig:system_overview}b). Five different poses were selected to reduce the likelihood of participants memorizing a unique pose and not paying attention to the Trainer's guidance.

The experiment took place at the Motor Learning and Neurorehabilitation (MLN) Laboratory in the Cognitive Robotics Department at TU Delft. Participants worked in dyads. At the beginning of each experiment session, members of the dyad were randomly assigned to the role of Trainer or Trainee and received instructions regarding the assigned tasks. They then completed an initial questionnaire on Qualtrics (Qualtrics, Provo, UT; version 2024), collecting demographic information 
and prior experience using robots or immersive virtual reality systems.

The Trainee was seated comfortably and assisted by the experimenter in donning the ARMin exoskeleton, while the Trainer was seated in a remote area of the laboratory and assisted in wearing the VR headset.
Each desired pose was first presented to the Trainer in the virtual environment, after which the Trainer was instructed to guide the Trainee toward the target pose as quickly as possible. Each target pose was preceded by a base pose, defined as a relaxed resting posture with the Trainee’s hand placed on a virtual table. 
Two experimental conditions were tested:

\begin{enumerate}
    \item \textbf{Visual Demonstration (VD):} The Trainee could observe the Trainer via a screen positioned in front of them. The Trainer demonstrated the arm poses with their own arms in front of the webcam, and the Trainee was verbally instructed to imitate the pose. Verbal communication was always permitted. 
    \item \textbf{Haptic Demonstration (HD):} The Trainer used the Sigma.7 to manipulate the ARMin digital twin and physically assist the Trainee in positioning their arm toward the correct pose. As in the previous condition, verbal communication was allowed to the Trainer, but this time, no visual clue was provided to the Trainee.
\end{enumerate}%
Concurrent visual feedback was provided exclusively to the Trainer during task execution. In particular, the target poses were represented by a light blue arm (see Fig. \ref{fig:system_overview}b), with two markers indicating the desired positions for the elbow and wrist. This arm turned green when the Trainee successfully reached and maintained the pose, i.e., when the wrist and elbow positions of the Trainee's avatar matched the positions of the target arm within \qty{7}{cm} and were maintained for at least 3 seconds. After reaching the desired pose, the Trainee was instructed to return to the base pose.

Each Trainer-Trainee dyad went through six task execution blocks, three per condition (VD and HD). Half of the dyads underwent the VD condition first, while the other half began with the HD condition. Each block included the five possible ADL poses, executed in random order, along with the required five repetitions to return to the base pose. Only paths towards the five poses were considered for analysis. Prior to the first block of each condition, participants completed a familiarization phase consisting of three random pose demonstrations. Blocks were interspersed with pauses of two minutes to reduce fatigue. Additionally, the ARMin exoskeleton provided assistance throughout the duration of the experiment in the form of arm weight compensation (~65\% of their arm weight was compensated following the work of Just \textit{et al.}~\cite{just2020human}).

After completion of each condition, participants filled out another questionnaire, including questions related to their intrinsic motivation, engagement, and perceived mental workload, as detailed in Sec.~\ref{sec: subjective questions}. At the end of the experiment, each dyad participated in an interview, where the experimenter collected general remarks and comments. Questions asked during the interviews are reported in the \textit{Supplementary Materials} of this paper. 
The experiment lasted approximately 90 minutes. 

\subsection{Outcome Metrics}
\subsubsection{Motion Outcomes --- Movement Quality and Completion Time}

For each pose execution, we recorded the ARMin joint angular positions with a sampling frequency of \SI{500}{\hertz}. We then derived the 3D positions of the graspable wrist and elbow points and calculated:
\begin{itemize}
    \item \textbf{Completion time}: the time Trainees needed to reach the instructed arm pose, with both elbow and wrist, from the moment a new pose is shown. We included the three maintenance seconds needed to consider the pose correctly executed. 
    The completion time gives an idea of how efficiently the Trainer can communicate the desired pose to the Trainee and how quickly the Trainee can interpret and execute the guidance.
    \item \textbf{Smoothness}: the smoothness of the graspable elbow and wrist movements, derived through the Spectral Arc length (SPARC)~\cite{ balasubramanian2015analysis}. The SPARC calculation requires the point's velocity, which was derived from the position data and filtered with a first-order low-pass Butterworth filter (\SI{20}{\hertz} cutoff frequency~\cite{ antonsson1985frequency}). The parameters employed in the SPARC calculation include a cutoff frequency of $ w_{max}$ = \SI{20}{\hertz}, and amplitude threshold $V$ = 0.05. SPARC is a negative measure, where lower absolute values indicate higher movement smoothness. Increased smoothness is commonly used in rehabilitation contexts to indicate well-directed movements, with minimal corrective submovements, and here reflects the quality of the instructed movements.
\end{itemize}
The completion time and smoothness calculations were performed using Python 3.8.10 (Python Software Foundation, USA).

\subsubsection{Verbal Interaction Analysis --- Total and Percentual Instructions Provided to Trainees}

We recorded (Philips, VoiceTracer, 8GB, Netherlands) the verbal communication provided by Trainers to their Trainees while they were performing the task under the two conditions. The speech recordings were automatically transcribed using the Whisper speech-to-text model (OpenAI)~\cite{radford2022whisper}, which was run locally on Windows 11. Audio preprocessing was handled through the FFmpeg framework~\cite{ffmpeg}, and sentences were isolated with the pre-trained Punkt tokenizer (NLTK)~\cite{kiss2006unsupervised}. After sentence splitting, each sentence was embedded into a vector representation using the all-MiniLM-L6-v2 model from the Sentence-BERT family~\cite{reimers2019}. 

The embeddings were then clustered with k-means. Each cluster was assigned a representative label corresponding to the most frequent sentence label within that cluster. The labels (INSTRUCTION, FEEDBACK, NULL) were defined through manual inspection of one representative recording, collected from the first dyad. Each sentence of this recording was categorized according to its communicative function, and the labels were used to train the clustering-based classifier. For instance, sentences such as ``Bend your elbow'' or ``Lift your arm'' were annotated as INSTRUCTION; expressions like ``Good'' or ``Perfect'' were labeled as FEEDBACK, and fillers or other expressions such as ``So'' or ``Sorry'' were assigned to the NULL class.

A second recording from another pair of participants was labeled and used for testing, allowing us to evaluate the model’s generalizability. We evaluated different options for the clustering number $k$, aiming for the best testing accuracy, and ultimately selected $k=10$, which yielded an accuracy of $75\%$.
All remaining recordings were processed automatically without manual inspection to minimize annotation effort.

Finally, we computed: (i) the total number of instructional sentences provided by the Trainer of each dyad, per trial and per condition, and (ii) the percentage of instructional sentences over the total of sentences extracted from each recording. The total number of instructions was used to quantify the absolute amount of verbal guidance required to complete the task, serving as an indicator of how much Trainers complemented their demonstration modality (visual or haptic) with verbal interactions. The percentage, instead, normalized this measure with respect to overall speech activity, allowing comparisons across conditions and dyads with different amounts of speech. Together, these measures enable us to assess whether the proposed haptic framework reduced the need for explicit verbal instructions as a complement to the demonstration modality, which could otherwise influence task completion time and movement smoothness.
The analysis focused on instructional sentences only, without further investigating feedback-related content. Nevertheless, the inclusion of both FEEDBACK and NULL labels during the clustering stage was adopted as a design choice to reduce ambiguity and, consequently, limit potential contamination of the INSTRUCTION class. 

\subsubsection{Subjective Outcomes --- Questionnaire}
\label{sec: subjective questions}
The following subjective outcome metrics were computed from the questionnaires filled in by all participants after completion of the last block, one for each condition:
\begin{itemize}
    \item \textbf{Motivation}: evaluated through two items from the Effort and Perceived Competence subscales of the Intrinsic Motivation Inventory (IMI)~\cite{mcauley1989psychometric}. Each item was responded to on a 7-point Likert scale, where 1 indicated ``not at all true,'' and 7 indicated ``very true.''
    \item \textbf{Mental workload}: assessed using the raw NASA Task Load Index (RTLX)~\cite{hart2006nasa}, which consists of six distinct domains: mental demand, physical demand, temporal demand (referring to perceived time pressure), perceived performance, effort (the amount of effort required to achieve the performance), and frustration level. Each domain was measured on a single 21-point Likert-style scale. In this scale, a score of zero indicates ``very low'', while a score of 20 indicates ``very high''~\cite{ratz2024designing}.
\end{itemize}

\subsection{Statistical Analysis}
Statistical analyses were performed using R (R Studio, version 2024.04.1). 
For each dyad, quantitative metrics were recorded along three blocks per condition, with five ADL poses per block. 
For each quantitative outcome metric, we excluded outliers using the criterion: $< Q1 - 2.0\cdot IQR$ or $>Q3 + 2.0\cdot IQR$, where IQR is the interquartile range, and $Q1$ and $Q3$ are the first and third quartiles, respectively. The use of a $2.0\cdot IQR$ threshold (rather than the conventional $1.5\cdot IQR$ fence) reduces excessive trimming in skewed or heavy-tailed distributions while still removing extreme values.

We employed linear mixed models (LMMs), implemented with the $lmerTest$ R package~\cite{kuznetsova2017lmertest}, to evaluate whether participants performed differently when executing the task under the two different demonstration \textbf{Conditions}. In all models, the \textbf{Block} number (coded 1-3 within each condition) was included to capture possible learning effects specific to each demonstration modality. \textbf{Condition order} (i.e., whether participants performed first the VD condition ---1--- or the HD condition ---2---) was added to assess potential order-related biases. To account for inter-individual variability, \textbf{Dyad} (corresponding to the dyad number) was modeled as a random effect.

\begin{align}
    \text{Outcome\_measure} \sim & \ \text{Condition} \times (\text{Block} + \text{Condition\_Order} ) + (1|\text{Dyad})
\end{align}

For all models, the intercept corresponds to the outcome measure in the VD condition, on the first block, with the VD condition presented first. All the effects are expressed as differences from this reference. 

To establish whether there were differences between the two conditions in the participants' perceived workload and motivation, the participants' RTLX and IMI scores were also evaluated using LMMs. RTLX scores from the individual categories were rescaled to be on a 0-100 scale. The \textbf{Role} of the participant (Trainer or Trainee), the \textbf{Condition} and their interaction were added as fixed effects. 
\textbf{Participant} (corresponding to a number per individual participant) was included as a random effect. The model is described by the following equation:
\begin{align}
\text{Questionnaire\_Score} &\sim \text{Role} \times \text{Condition} + (1|\text{Participant}).
\end{align}

Post hoc analyses were performed on the results of all the aforementioned models using FDR correction. 
The significance level chosen to interpret results was set at $\alpha = 0.05$ for all statistical tests.

Lastly, we fitted an LMM on the total and percentage numbers of instructions provided by Trainers to Trainees. In particular, we investigated \textbf{Condition} and \textbf{Block} as fixed effects, and Dyad as a random effect. 

\begin{align}
\text{Instructions\_(Tot/Perc)} &\sim \text{Condition} + \text{Block} + (1|\text{Dyad})
\end{align}

Note that for the evaluation of questionnaire scores and verbal instructions, we did not include \textbf{Condition order} in the models, as we did not expect that learning or fatigue would influence these variables.
For the verbal‑interaction analysis, we similarly did not investigate interaction effects, as our interest was limited to assessing overall differences between the two demonstration modalities. \text{Block} was included only as a simple covariate to capture potential general trends over time, rather than as a factor of substantive interest.

\section{Results}
In total, we recorded 480 trials (16 pairs of participants $\times$ 5 poses $\times$ 3 blocks $\times$ 2 conditions). Outlier removal and the resulting counts and percentages per metric and condition are reported in the \textit{Supplementary Material}.

\subsection{Movement Quality and Completion Time}
\subsubsection{Completion time}
Our linear mixed model results (Tab.~\ref{tab: time}) indicated an effect of \textbf{Condition} on completion time, with the HD condition associated with significantly shorter times than the VD ($p_{FDR} = 0.018$).
We did not find a main effect of the \textbf{Block} or \textbf{Condition order}, nor a significant interaction between \textbf{Condition} and \textbf{Block}. However, we found a significant interaction between \textbf{Condition} and \textbf{Condition order} ($p_{FDR} = 0.005$), with the HD condition producing faster completion times if the VD was presented first.

\begin{table}[h!] 
\centering 
\resizebox{0.70\textwidth}{!}{ 
\begin{tabular}{lcccccccc} 
\toprule
 & Estimate & Std. Error & 95\% CI & df & t value & Pr($>|t|$) & $p_{FDR}$ \\ 
\midrule
(Intercept) & 20.81 & 1.90 & [16.87, 24.75] & 22.35 & 11.00 & 1.72e-10 & 1.03e-09*** \\
HD          & -3.84 & 1.46 & [-6.71, -0.97] & 439.29 & -2.63 & 0.009 & 0.018* \\
Block       & -0.70 & 0.66 & [-2.00, 0.60]  & 439.10 & -1.07 & 0.287 & 0.430\\
HD\_first      & 0.65 & 2.50 & [-4.63, 5.93]  & 16.97  & 0.26 & 0.798 & 0.798\\
HD:HD\_first   & 4.88  & 1.53 & [1.87, 7.89]   & 439.17 & 3.19  & 0.002 & 0.005** \\
HD:Block    & -0.62 & 0.95 & [-2.49, 1.25]  & 439.24 & -0.66 & 0.509 & 0.611 \\
\bottomrule 
\end{tabular} 
} 
\caption{Results from the LMM with Completion Time. CI: Wald 95\% confidence intervals. Condition Order:2 when the HD condition is the first. For clarity, in the table, we reported ``HD\_first'' instead of Condition Order. .($p < 0.1$), *($p<0.05$), **($p<0.01$), ***($p<0.001$).} 
\label{tab: time} 
\end{table}

\subsubsection{Smoothness}
Results from the linear mixed model for elbow and wrist smoothness (Tab.~\ref{tab: smooth}) indicated that the HD condition significantly improved elbow smoothness ($p_{FDR}=0.046$), while improvements at the wrist did not approach significance ($p_{FDR}=0.207$). 

The \textbf{Block} main effect also showed improvements in smoothness at both joints, albeit non-significant.
The interaction between \textbf{Condition} and \textbf{Condition order} for both joints indicated that poses were executed with poorer smoothness (lower values) when HD was presented first ($p_{FDR, elbow}=0.007$, $p_{FDR, wrist}=0.001$). 

\begin{table}[h!]
\centering
\resizebox{0.95\textwidth}{!}{
\begin{tabular}{lccccccc|ccccccc}
\toprule
& \multicolumn{7}{c}{\textbf{Elbow SPARC [-]}} & \multicolumn{7}{c}{\textbf{Wrist SPARC [-]}} \\ 
\cmidrule(lr){2-8} \cmidrule(lr){9-15}
 & Est. & Std. Er. & 95\% CI & df & t value & Pr($>|t|$) & $p_{\mathrm{FDR}}$ 
 & Est. & Std. Er. & 95\% CI & df & t value & Pr($>|t|$) & $p_{\mathrm{FDR}}$ \\ 
\midrule
(Intercept)       
 & -4.20 & 0.34 & [-4.87, -3.53] & 28.01 & -12.36 & 7.40e-13 & 4.44e-12*** 
 & -3.92 & 0.33 & [-4.57, -3.27] & 25.08 & -11.78 & 1.02e-11 & 6.11e-11*** \\

HD            
 & 0.70 & 0.31 & [0.09, 1.31] & 450.15 & 2.28 & 0.023 & 0.046* 
 & 0.54 & 0.30 & [-0.05, 1.13] & 447.10 & 1.95 & 0.052 & 0.104 \\

HD\_first            
 & 0.05 & 0.41 & [-0.75, 0.85] & 19.43 & 0.11 & 0.912 & 0.966 
 & 0.19 & 0.41 & [-0.61, 1.00] & 18.31 & 0.45 & 0.654 & 0.669 \\

Block             
 & 0.25 & 0.14 & [-0.02, 0.52] & 450.16 & 1.80 & 0.073 & 0.110 
 & 0.15 & 0.12 & [-0.09, 0.39] & 447.19 & 1.19 & 0.235 & 0.353 \\

HD:HD\_first     
 & -0.97 & 0.32 & [-1.60, -0.34] & 450.11 & -3.04 & 0.003 & 0.007* 
 & -1.04 & 0.29 & [-1.61, -0.47] & 447.08 & -3.61 & 3.41e-04 & 0.001** \\

HD:Block      
 & 0.008 & 0.19 & [-0.36, 0.38] & 450.13 & 0.04 & 0.966 & 0.966 
 & 0.07 & 0.18 & [-0.28, 0.42] & 447.09 & 0.43 & 0.669 & 0.669 \\
\bottomrule
\end{tabular}
}
\caption{Results of the LMMs for smoothness ---evaluated through SPARC--- at elbow and wrist graspable points. CI: Wald 95\% confidence intervals. Condition Order:2 when the HD condition is the first. For clarity, in the table, we reported ``HD\_first'' instead of Condition Order. .($p < 0.1$), *($p < 0.05$), **($p < 0.01$), ***($p < 0.001$).}
\label{tab: smooth}
\end{table}

\subsection{Verbal Interaction Analysis}
\begin{figure}[h!]
    \centering
\includegraphics[width=0.7\linewidth]{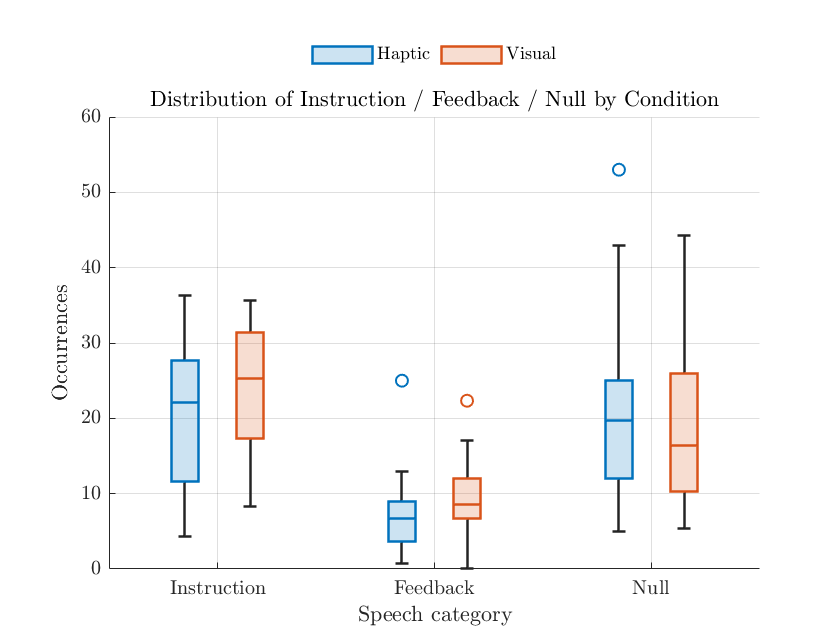}
    \caption{Summary of the results of the speech analysis, with numbers of occurrences for Instruction, Feedback, and Null under Haptic (blue) and Visual (red) conditions. The boxes show the medians and interquartile ranges. }
    \label{fig:speech}
\end{figure}

Fig.~\ref{fig:speech} represents the results of the speech analysis in terms of the number of sentences, per label, said by the Trainers in the two conditions. From these results, we isolated the number of sentences labeled as INSTRUCTIONS and evaluated their percentage with respect to the total number of sentences. In the VD condition, Trainers generally provided more instructions than in HD mode, although the difference did not reach significance (Tab.~\ref{tab:speech}). However, we found that the total number of instructions significantly decreased across blocks ($p = 0.003$). %
The percentage of instructions over the total number of sentences followed the contrary pattern, i.e., we found a significant decrease in the percentage of instructions in the HD condition compared to VD ($p = 0.017$), while the effect of Block was not significant.

\
\begin{table}[h!]
    \centering
    \resizebox{0.7\textwidth}{!}{%
\begin{tabular}{lcccccc}
\toprule
\multicolumn{7}{c}{\textbf{Total Instructions}} \\ 
\midrule
 & Estimate & Std. Error & 95\% CI & df & t value & Pr($>|t|$) \\ 
\midrule
(Intercept) & 32.81 & 3.59 & [25.69, 39.94] & 99.89 & 9.14 & 7.67e-15 *** \\
HD      & -3.22  &  2.37 & [-7.93, 1.48] & 87.15 & -1.36 & 0.177 \\
Block       & -4.47 & 1.45 & [-7.37, -1.58] & 87.23 & -3.08 & 0.003 ** \\
\midrule
\multicolumn{7}{c}{\textbf{Percentage Instructions}} \\ 
\midrule
(Intercept) & 53.59  &   3.94 & [45.77, 61.40] & 103.66 & 13.61 & $<$2e-16 *** \\
HD      & -6.56  &  2.69 & [-11.91, -1.21] & 87.40 & -2.44 & 0.017 * \\
Block       & -1.03 & 1.65 & [-4.31, 2.25] & 87.49 & -0.63 & 0.533 \\
\bottomrule
\end{tabular}
}
\caption{Results of the LLMs for Total Instructions and Percentage of Instructions provided by Trainers to Trainees. CI: Wald 95\% confidence intervals. *($p <0.05$), **($p <0.01$), ***($p <0.001$).}
\label{tab:speech}
\end{table}

\subsection{Questionnaire Outcomes}
\subsubsection{Mental Workload}
\begin{figure}[h!]
    \centering
\includegraphics[width=\linewidth]{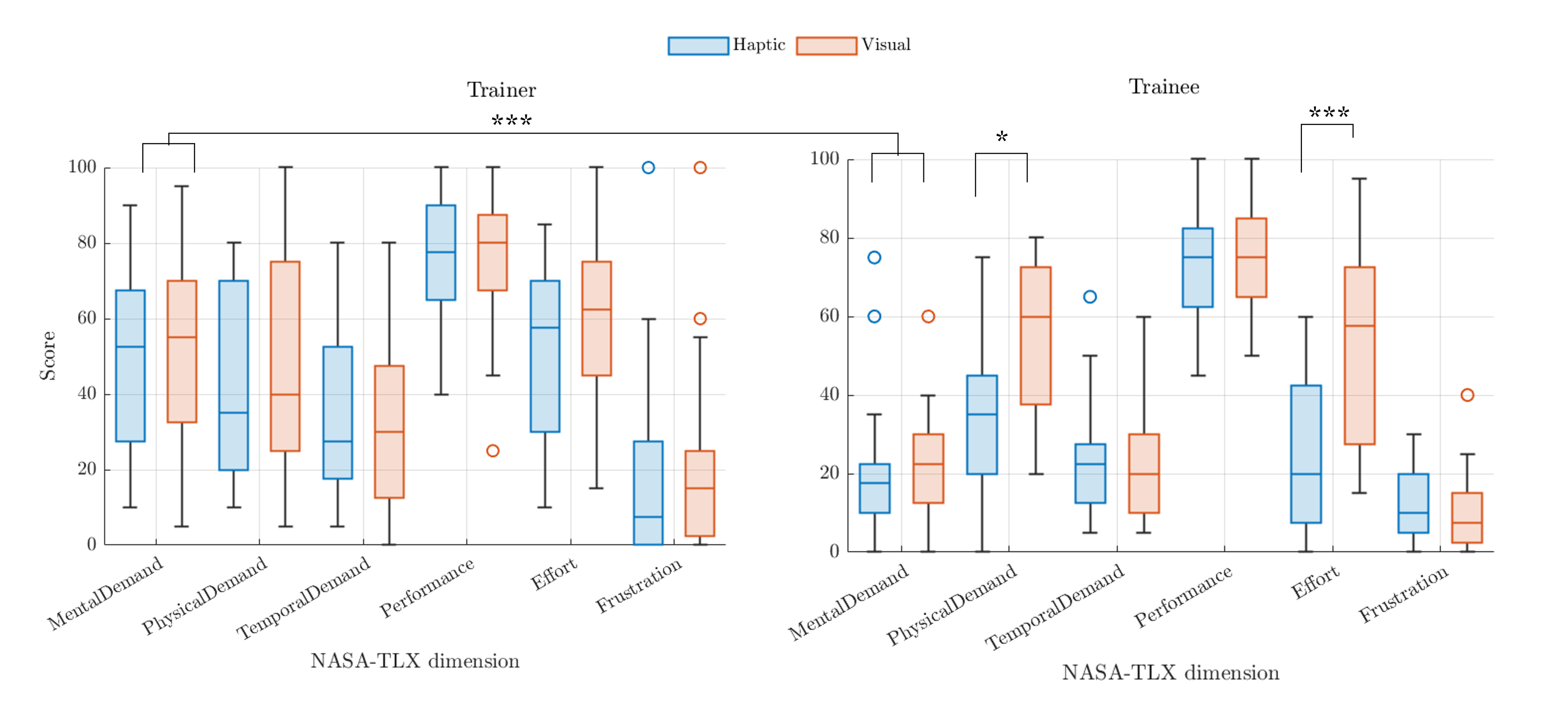}
    \caption{Summary of the RTLX scores, divided per dimension and role within the dyad. The boxes show the medians and interquartile ranges. Significant differences, based on the results of the LLMs, are reported: *(p < 0.05), **(p < 0.01), ***(p < 0.001).}
    \label{Fig: Nasa}
\end{figure}

Fig.~\ref{Fig: Nasa} presents the summary of the RTLX  scores for each subscale from participants in the roles of Trainers and Trainees. 
The full LMM analysis is reported in the \textit{Supplementary Materials}. Here, we only comment on the most relevant results for the various dimensions:
\begin{itemize}
    \item Participants' \textbf{Role} affected the reported \textit{Mental Demand}. In particular, Trainers reported significantly higher mental demand than Trainees ($p_{FDR} < 0.001$). The moderate mental demand reported was not significantly further reduced with the use of haptic demonstration.
    \item The HD condition was associated with lower \textit{Physical Demand} than VD for the Trainees group ($p_{FDR} = 0.034$).
    Being a Trainer or a Trainee did not significantly alter the perception of \textit{Physical Demand} ($p_{FDR} > 0.1 $). 
    \item The HD condition was also associated with significantly lower \textit{Effort} than the VD for Trainees ($p_{FDR} = 0.001$), with no significant difference in the Trainers' Role. 
    \item For the other dimensions, namely \textit{Frustration}, \textit{Temporal Demand}, and \textit{Performance}, no significant main or interaction effects were found.
\end{itemize}

\subsubsection{Intrinsic Motivation}
The summary of the reported scores for the Effort and Perceived Competence subscales of the IMI is shown in Fig.~\ref{fig: motivation}, and the related LMMs analysis is reported in the \textit{Supplementary Materials}. 
\begin{figure}[h!]
    \centering
\includegraphics[width=\textwidth]{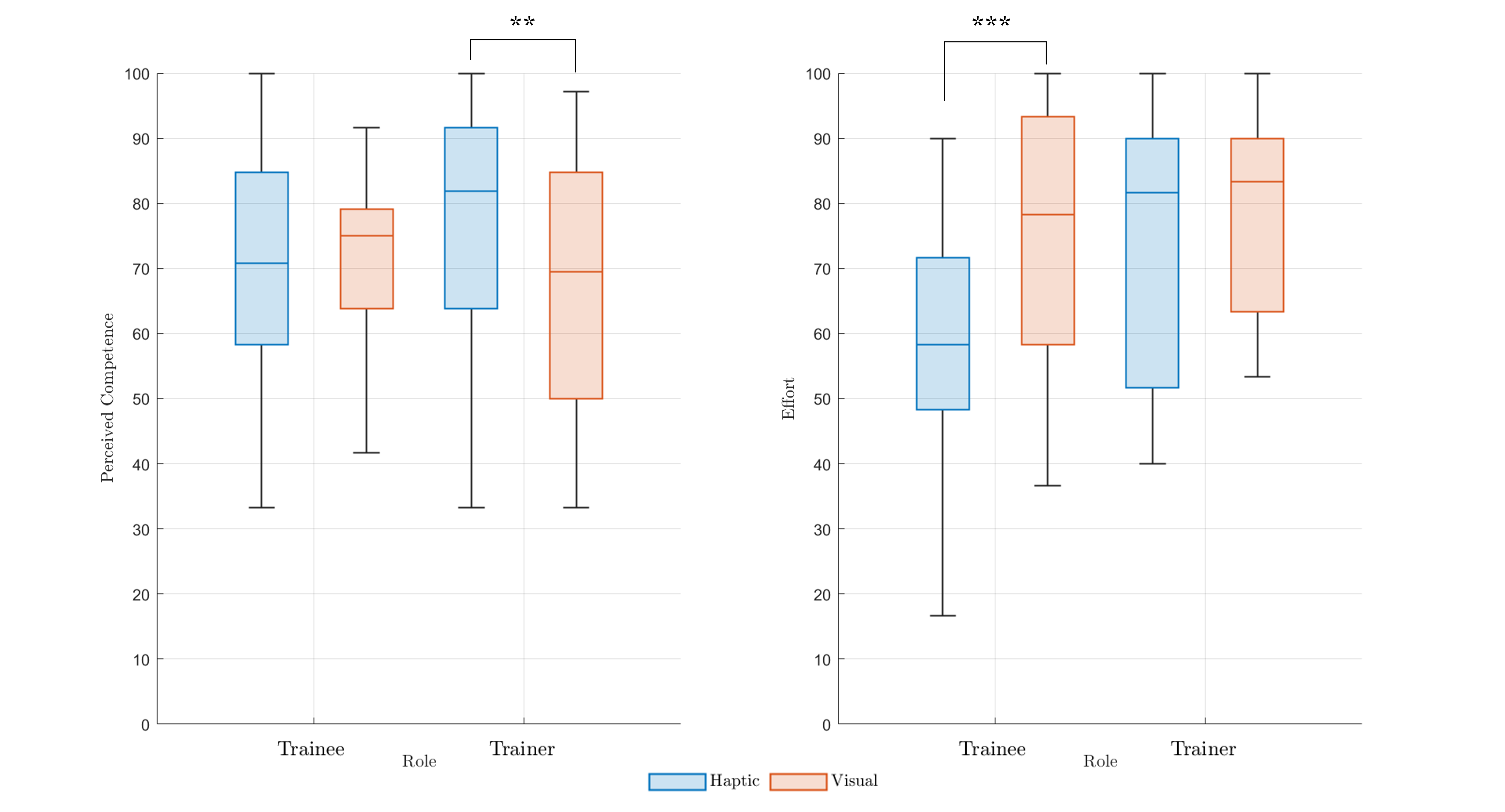}
    \caption{Average Effort and Perceived Competence subscales of the IMI. Significant differences, based on the results of the LLMs, are reported: *(p < 0.05), **(p < 0.01), ***(p < 0.001).}
    \label{fig: motivation}
\end{figure}
The interaction between \textbf{Role} and \textbf{Condition} affected the Perceived Competence sub-category. In particular, Trainers perceived themselves as more competent in the HD condition compared to the VD condition ($p_{FDR} = 0.004$). When analyzing the Effort subcategory, we found both a main effect of \textbf{Condition} ($p_{FDR}<0.001$) and an interaction between \textbf{Role} and \textbf{Condition} ($p_{FDR} = 0.020$). In particular, Trainees reported perceiving less effort in the HD condition compared to the VD condition, and Trainers reported higher effort than Trainees in the HD condition.

\section{Discussions}

We presented a novel teleoperation haptic therapist-in-the-loop framework that provides trainers with a fast and effective means of interacting with trainees while using rehabilitation robots. Our approach enables Trainers/Therapists to utilize commercial, easy-to-use end-effector haptic devices to physically guide the Trainees/Patients' arms while wearing supporting exoskeletons. Below, we discuss the main findings of our feasibility study.

\subsection{The Trainers Could Physically Adjust Trainees' Upper-limb Poses More Effectively than with Visual Demonstration }

Using our haptic therapist-in-the-loop system allowed the Trainers to guide Trainees more efficiently compared to a more conventional visual demonstration approach. Movements were completed significantly faster, with an average reduction of approximately \SI{3.9}{\second} per pose, and trajectories---especially at the elbow---were smoother. These results indicate that remote haptic cues provided a more immediate and precise mean of correcting the Trainee’s motion. Importantly, this more efficient HD interaction did not come at the cost of higher Trainers' mental/physical demand or higher frustration, suggesting that the interaction was indeed intuitive.  

The smoother movements executed within the HD condition, however, could also be attributed to the use of the teleoperation controller on top of the ARMin's baseline controller, which may have stabilized Trainees’ movements. Indeed, participants often reported in the post-experiment interviews that the zero-torque controller of the ARMin, when used alone, felt sometimes unstable and required time to adapt to, in line with the findings of Dalla Gasperina \textit{et al.}~\cite{dalla2023quantitative}. 
In both conditions (HD and VD), completion times decreased and smoothness increased across trials, although non-significantly, suggesting that participants indeed learned or adapted to the system. Participants who experienced VD first completed the subsequent HD trials more quickly, suggesting that an initial exposure to the use of the exoskeleton helped the dyads adapt to the introduction of a second robotic interface before effectively exploiting the haptic modality. This is coherent with the fact that, in human-robot interactions, humans often require some exposure to the robotic system to familiarize themselves before adjusting their abilities and achieving consistent performance~\cite{Poggensee2021}. 

The most efficient interaction between dyad members during HD is supported by the results from their verbal communication evaluation. Participants in the Trainer role needed  fewer verbal instructions to communicate the final pose when using our haptic therapist-in-the-loop device vs the visual demonstration. This effect can indicate that physical guidance serves as an efficient implicit, sensorimotor channel to transmit task-related information from therapists to patients, preserving aspects of the physical human-human interactions that are fundamental in conventional rehabilitation~\cite{sawers2014}, and that might not be equally conveyed with visual demonstrations.
When we normalized the number of instructions to the total number of verbal exchanges, it became clear that the haptic demonstration significantly reduced the proportion of instructions. This suggests a selective reduction in instructional content, while overall verbal engagement remained similar across conditions. Trainers did not talk less overall, but the content of their speech changed. This could be an important aspect during therapy since, as demonstrated in a recent study by Di Tommaso \textit{et al.}~\cite{DiTommaso2025}, during robot-assisted training, therapists' verbal interactions can improve active participation in patients. 
By reducing the reliance on verbal cues for conveying task-related guidance, our system could allow the therapist to focus their verbal interactions on other crucial communicative functions---such as encouragement, feedback, or emotional support---which are important facilitators in rehabilitation~\cite{Michielsen14082019, Cucinella2024}. 

These encouraging words might be especially relevant when physically guiding patients' movements. In line with the Guidance Hypothesis, we found that Trainees reported significantly lower perceived effort and physical demand in the HD vs VD condition. When moving to clinical settings, a lack of patient involvement can be detrimental~\cite{Woodbridge2023}, and it may lead to the so-called slacking effect, where patients decrease their voluntary effort due to receiving excessive assistance, ultimately limiting motor learning and recovery~\cite{Basalp2021}. 
An important clinical advantage of our approach, however, lies in the therapist’s ability to decide \textit{when} during the patient's movement execution the haptic support is needed. With our teleoperation system, they can intervene and give support when they feel it is necessary and/or only in specific moments. Given their knowledge of the patient’s condition--- which they can nevertheless also feel through the end-effector device as forces are projected back to the haptic device---therapists can be trusted not to take over the patient’s effort, thus mitigating the risk of slacking.

Taken together, these considerations indicate that the implementation of our teleoperation controller was successful, and Trainers could efficiently transmit their inputs to the Trainees without the need for high physical strain and while achieving high levels of subjective satisfaction. While previous systems often restricted interaction to the end‑effector~\cite{hou2025} or required matching exoskeletons on both sides~\cite{baur2019beam}, our results show that a single end-effector haptic interface can be used to provide sufficient control and manipulability for intuitive multi-joint remote physical.

\subsection{Clinical Implications}
The proposed framework has relevant practical implications for robot-assisted rehabilitation. 
Through the use of our therapist-in-the-loop haptic system, we reintroduce the therapist's physical touch in robotic rehabilitation. In this way, we restore the movement-facilitator~\cite{Michielsen14082019}, an essential component of conventional rehabilitation that therapists lose in robot-assisted therapy, with exoskeletons especially~\cite{LucianiTAM}. By doing so, we can reduce the common perception therapists have of robots as barriers, rather than supports~\cite{Lo2020}, potentially enhancing their acceptability~\cite{LucianiTAM}. 
By enabling joint-level control of patients' movements, our system could allow therapists to promptly correct postures, avoid compensation strategies, or stabilize movements. In this way, we could extend the personalization options offered by robots as exoskeletons to address the needs of patients with different levels of not only motor but also cognitive impairment, for whom physical cues may be more intuitive than complex verbal or visual instructions. 

\subsection{Limitations and Future Developments}
Our study suffers from some limitations. First, we included Trainees who were both left- and right-handed individuals. Yet, they were all tested using the right-handed ARMin configuration. To limit the complexity of our statistical analysis, we did not include handedness as a factor in our statistical model. Yet, we do not expect this to influence our results, as Trainees performed both conditions with the same hand. 

Secondly, we tested the feasibility of our system only with healthy participants with technological experience, but without a clinical background. To properly assess its potential as a support for therapists, we need to investigate therapists' perception, trust, and system usability. 
One of the aspects worth investigating concerns the use of the headset. Immersive VR reduces the user's cognitive load and promotes more efficient 3D movements when compared to screens~\cite{wenk}, making it an attractive tool in telerehabilitation contexts. However, we should involve therapists in the evaluation of other display alternatives (e.g., augmented reality) in clinical settings, where the VR headset might limit the perception of the surrounding environment, potentially reducing therapists' ability to assess some aspects of patients' needs \cite{Brady2023}. 
The system should also be evaluated with patients, not only to assess usability but also to identify which subgroups of patients would benefit the most from this kind of support, and in which therapeutic conditions.

\section{Conclusions}
This study introduces a novel therapist-in-the-loop haptic teleoperation system to restore joint-level therapist-patient physical interaction during robot-assisted rehabilitation. By enabling therapists to use a commercially available end-effector haptic device to remotely guide patients in an upper-body exoskeleton through haptic contact points at the elbow and wrist, the system addresses a well-known limitation of current rehabilitation robots: the loss of human-human physical contact.
We conducted a feasibility study with healthy individuals, in dyads, to assess how our system can enhance the capabilities of Trainers to guide Trainees' movements. Results indicate that the use of our system improved task performance significantly more than conventional visual demonstrations.
Movements were completed more quickly and smoothly, and Trainers relied less on verbal instructions to convey task-related information, suggesting a more efficient sensorimotor communication. Importantly, Trainers reported a greater sense of competence without increased mental and physical workload.  Trainees experienced reduced physical effort when being physically guided, highlighting the importance of balancing Trainers' assistance with Trainees' engagement in future clinical use.
Our framework has the potential to reintroduce therapeutic touch into robot‑assisted rehabilitation and, by doing so, to improve therapists’ acceptance of robotic technologies. Future work should evaluate the system with clinicians and neurological patients to assess usability, therapeutic impact, and the role of patient‑specific factors---such as motor impairment level---on treatment effectiveness, to define the optimal target population for this technology.

\bibliography{files/biblio}

\begin{thebibliography}{10}
\urlstyle{rm}
\expandafter\ifx\csname url\endcsname\relax
  \def\url#1{\texttt{#1}}\fi
\expandafter\ifx\csname urlprefix\endcsname\relax\def\urlprefix{URL }\fi
\expandafter\ifx\csname doiprefix\endcsname\relax\def\doiprefix{DOI: }\fi
\providecommand{\bibinfo}[2]{#2}
\providecommand{\eprint}[2][]{\url{#2}}

\bibitem{Mehrholz08}
\bibinfo{author}{Mehrholz, J.}, \bibinfo{author}{Pohl, M.}, \bibinfo{author}{Platz, T.}, \bibinfo{author}{Kugler, J.} \& \bibinfo{author}{Elsner, B.}
\newblock \bibinfo{journal}{\bibinfo{title}{Electromechanical and robot‐assisted arm training for improving activities of daily living, arm function, and arm muscle strength after stroke}}.
\newblock {\emph{\JournalTitle{Cochrane Database of Systematic Reviews}}}  (\bibinfo{year}{2018}).

\bibitem{MarchalCrespo2009}
\bibinfo{author}{Marchal‐Crespo, L.} \& \bibinfo{author}{Reinkensmeyer, D.~J.}
\newblock \bibinfo{journal}{\bibinfo{title}{Review of control strategies for robotic movement training after neurologic injury}}.
\newblock {\emph{\JournalTitle{Journal of NeuroEngineering and Rehabilitation}}} \textbf{\bibinfo{volume}{6}}, \bibinfo{pages}{20} (\bibinfo{year}{2009}).

\bibitem{Longatelli2023}
\bibinfo{author}{Longatelli, V.}, \bibinfo{author}{Luciani, B.}, \bibinfo{author}{Pedrocchi, A.} \& \bibinfo{author}{Gandolla, M.}
\newblock \bibinfo{title}{Instrumented upper limb functional assessment using a robotic exoskeleton: Normative references intervals}.
\newblock In \emph{\bibinfo{booktitle}{2023 International Conference on Rehabilitation Robotics (ICORR)}}, \bibinfo{pages}{1--6} (\bibinfo{year}{2023}).

\bibitem{Richard2020}
\bibinfo{author}{Richard, E.}, \bibinfo{author}{Woolsey, C.}, \bibinfo{author}{McGibbon, C.} \& \bibinfo{author}{O'Connell, C.}
\newblock \bibinfo{journal}{\bibinfo{title}{Physiotherapists’ experiences using the ekso bionic exoskeleton with patients in a neurological rehabilitation hospital: A qualitative study}}.
\newblock {\emph{\JournalTitle{Rehabilitation Research and Practice}}} \textbf{\bibinfo{volume}{2020}}, \bibinfo{pages}{1--8} (\bibinfo{year}{2020}).

\bibitem{Lo2020}
\bibinfo{author}{Lo, K.}, \bibinfo{author}{Stephenson, M.} \& \bibinfo{author}{Lockwood, C.}
\newblock \bibinfo{journal}{\bibinfo{title}{Adoption of robotic stroke rehabilitation into clinical settings: a qualitative descriptive analysis}}.
\newblock {\emph{\JournalTitle{International Journal of Evidence-Based Healthcare}}} \textbf{\bibinfo{volume}{4}}, \bibinfo{pages}{376--390} (\bibinfo{year}{2020}).

\bibitem{LucianiTAM}
\bibinfo{author}{Luciani, B.}, \bibinfo{author}{Braghin, F.}, \bibinfo{author}{Pedrocchi, A. L.~G.} \& \bibinfo{author}{Gandolla, M.}
\newblock \bibinfo{journal}{\bibinfo{title}{Technology acceptance model for exoskeletons for rehabilitation of the upper limbs from therapists' perspectives}}.
\newblock {\emph{\JournalTitle{Sensors}}} \textbf{\bibinfo{volume}{23}} (\bibinfo{year}{2023}).

\bibitem{Platz2021}
\bibinfo{author}{Platz, T.}, \bibinfo{author}{Schmuck, L.}, \bibinfo{author}{Roschka, S.} \& \bibinfo{author}{Burridge, J.}
\newblock \bibinfo{title}{Arm rehabilitation}.
\newblock In \bibinfo{editor}{Platz, T.} (ed.) \emph{\bibinfo{booktitle}{Clinical Pathways in Stroke Rehabilitation: Evidence-based Clinical Practice Recommendations}}, \bibinfo{pages}{97--121} (\bibinfo{publisher}{Springer International Publishing}, \bibinfo{address}{Cham}, \bibinfo{year}{2021}).

\bibitem{Hasson2023}
\bibinfo{author}{Hasson, C.}, \bibinfo{author}{Manczurowsky, J.}, \bibinfo{author}{Collins, E.} \& \bibinfo{author}{Yarossi, M.}
\newblock \bibinfo{journal}{\bibinfo{title}{Neurorehabilitation robotics: how much control should therapists have?}}
\newblock {\emph{\JournalTitle{Frontiers in Human Neuroscience}}} \textbf{\bibinfo{volume}{17}} (\bibinfo{year}{2023}).

\bibitem{Molle2025}
\bibinfo{author}{Molle, R.} \emph{et~al.}
\newblock \bibinfo{journal}{\bibinfo{title}{Exploring priority parameters in physiotherapist decision models for tailoring robot-aided rehabilitation}}.
\newblock {\emph{\JournalTitle{International Journal of Social Robotics}}} \bibinfo{pages}{1--16} (\bibinfo{year}{2025}).

\bibitem{Stephenson2017}
\bibinfo{author}{Stephenson, A.} \& \bibinfo{author}{Stephens, J.}
\newblock \bibinfo{journal}{\bibinfo{title}{An exploration of physiotherapists' experiences of robotic therapy in upper limb rehabilitation within a stroke rehabilitation centre}}.
\newblock {\emph{\JournalTitle{Disability and rehabilitation. Assistive technology}}} \textbf{\bibinfo{volume}{13}}, \bibinfo{pages}{1--8} (\bibinfo{year}{2017}).

\bibitem{Celian2021}
\bibinfo{author}{Celian, C.} \emph{et~al.}
\newblock \bibinfo{journal}{\bibinfo{title}{{A day in the life: a qualitative study of clinical decision-making and uptake of neurorehabilitation technology}}}.
\newblock {\emph{\JournalTitle{Journal of NeuroEngineering and Rehabilitation}}} \textbf{\bibinfo{volume}{18}}, \bibinfo{pages}{121} (\bibinfo{year}{2021}).

\bibitem{Sommerhalder2022}
\bibinfo{author}{Sommerhalder, M.}, \bibinfo{author}{Kurth, N.}, \bibinfo{author}{Song, J.} \& \bibinfo{author}{Riener, R.}
\newblock \bibinfo{title}{Armstick - an intuitive therapist interface for upper-limb rehabilitation robots}.
\newblock In \emph{\bibinfo{booktitle}{2022 International Conference on Rehabilitation Robotics (ICORR)}}, \bibinfo{pages}{1--6} (\bibinfo{year}{2022}).

\bibitem{Pavón-Pulido2020}
\bibinfo{author}{Pavón-Pulido, N.}, \bibinfo{author}{López-Riquelme, J.} \& \bibinfo{author}{Feliú-Batlle, J.}
\newblock \bibinfo{journal}{\bibinfo{title}{Iot architecture for smart control of an exoskeleton robot in rehabilitation by using a natural user interface based on gestures}}.
\newblock {\emph{\JournalTitle{Journal of Medical Systems}}} \textbf{\bibinfo{volume}{44}} (\bibinfo{year}{2020}).

\bibitem{Amato2024}
\bibinfo{author}{Amato, L.} \emph{et~al.}
\newblock \bibinfo{title}{Unidirectional human-robot-human physical interaction for gait training} (\bibinfo{year}{2024}).
\newblock \eprint{https://arxiv.org/abs/2409.11510}.

\bibitem{Luciani2023}
\bibinfo{author}{Luciani, B.}, \bibinfo{author}{Roveda, L.}, \bibinfo{author}{Braghin, F.}, \bibinfo{author}{Pedrocchi, A.} \& \bibinfo{author}{Gandolla, M.}
\newblock \bibinfo{journal}{\bibinfo{title}{Trajectory learning by therapists' demonstrations for an upper limb rehabilitation exoskeleton}}.
\newblock {\emph{\JournalTitle{IEEE Robotics and Automation Letters}}} \textbf{\bibinfo{volume}{8}}, \bibinfo{pages}{4561--4568} (\bibinfo{year}{2023}).

\bibitem{Vianello2025}
\bibinfo{author}{Vianello, L.} \emph{et~al.}
\newblock \bibinfo{journal}{\bibinfo{title}{Robot-mediated physical human–human interaction in rehabilitation: A position paper}}.
\newblock {\emph{\JournalTitle{IEEE Reviews in Biomedical Engineering}}} \textbf{\bibinfo{volume}{19}}, \bibinfo{pages}{267--282} (\bibinfo{year}{2026}).

\bibitem{hou2025}
\bibinfo{author}{Hou, Z.} \emph{et~al.}
\newblock \bibinfo{title}{Learning a shape-adaptive assist-as-needed rehabilitation policy from therapist-informed input} (\bibinfo{year}{2025}).
\newblock \eprint{https://arxiv.org/abs/2510.04666}.

\bibitem{baur2019beam}
\bibinfo{author}{Baur, K.}, \bibinfo{author}{Rohrbach, N.}, \bibinfo{author}{Hermsd{\"o}rfer, J.}, \bibinfo{author}{Riener, R.} \& \bibinfo{author}{Klamroth-Marganska, V.}
\newblock \bibinfo{journal}{\bibinfo{title}{The “beam-me-in strategy”--remote haptic therapist-patient interaction with two exoskeletons for stroke therapy}}.
\newblock {\emph{\JournalTitle{Journal of neuroengineering and rehabilitation}}} \textbf{\bibinfo{volume}{16}}, \bibinfo{pages}{1--15} (\bibinfo{year}{2019}).

\bibitem{ratschat2023}
\bibinfo{author}{Ratschat, A.}, \bibinfo{author}{Lomba, T.~M.}, \bibinfo{author}{Dalla~Gasperina, S.} \& \bibinfo{author}{Marchal-Crespo, L.}
\newblock \bibinfo{title}{Development and validation of a kinematically accurate upper-limb exoskeleton digital twin for stroke rehabilitation}.
\newblock In \emph{\bibinfo{booktitle}{2023 International Conference on Rehabilitation Robotics (ICORR)}}, \bibinfo{pages}{1--6} (\bibinfo{organization}{IEEE}, \bibinfo{year}{2023}).

\bibitem{wenk}
\bibinfo{author}{Wenk, N.}, \bibinfo{author}{Buetler, K.}, \bibinfo{author}{Penalver, J.}, \bibinfo{author}{Müri, R.} \& \bibinfo{author}{Marchal-Crespo, L.}
\newblock \bibinfo{journal}{\bibinfo{title}{Naturalistic visualization of reaching movements using head-mounted displays improves movement quality compared to conventional computer screens and proves high usability}}.
\newblock {\emph{\JournalTitle{Journal of NeuroEngineering and Rehabilitation}}} \textbf{\bibinfo{volume}{19}} (\bibinfo{year}{2022}).

\bibitem{wenk2023effect}
\bibinfo{author}{Wenk, N.} \emph{et~al.}
\newblock \bibinfo{journal}{\bibinfo{title}{Effect of immersive visualization technologies on cognitive load, motivation, usability, and embodiment}}.
\newblock {\emph{\JournalTitle{Virtual Reality}}} \textbf{\bibinfo{volume}{27}}, \bibinfo{pages}{307--331} (\bibinfo{year}{2023}).

\bibitem{Christensen2023}
\bibinfo{author}{Christensen, N.}, \bibinfo{author}{Black, L.}, \bibinfo{author}{Gilliland, S.}, \bibinfo{author}{Huhn, K.} \& \bibinfo{author}{Wainwright, S.}
\newblock \bibinfo{journal}{\bibinfo{title}{The role of movement in physical therapist clinical reasoning}}.
\newblock {\emph{\JournalTitle{Physical Therapy}}} \textbf{\bibinfo{volume}{103}}, \bibinfo{pages}{pzad085} (\bibinfo{year}{2023}).

\bibitem{forcedimension}
\bibinfo{title}{Force dimension sigma.7 user manual}.
\newblock \bibinfo{note}{\url{https://www.forcedimension.com/products/sigma}}.

\bibitem{just2018exoskeleton}
\bibinfo{author}{Just, F.} \emph{et~al.}
\newblock \bibinfo{journal}{\bibinfo{title}{Exoskeleton transparency: feed-forward compensation vs. disturbance observer}}.
\newblock {\emph{\JournalTitle{at-Automatisierungstechnik}}} \textbf{\bibinfo{volume}{66}}, \bibinfo{pages}{1014--1026} (\bibinfo{year}{2018}).

\bibitem{dalla2023quantitative}
\bibinfo{author}{Dalla~Gasperina, S.}, \bibinfo{author}{Ratschat, A.~L.} \& \bibinfo{author}{Marchal-Crespo, L.}
\newblock \bibinfo{title}{Quantitative and qualitative evaluation of exoskeleton transparency controllers for upper-limb neurorehabilitation}.
\newblock In \emph{\bibinfo{booktitle}{2023 International Conference on Rehabilitation Robotics (ICORR)}}, \bibinfo{pages}{1--6} (\bibinfo{organization}{IEEE}, \bibinfo{year}{2023}).

\bibitem{schwarz2022characterization}
\bibinfo{author}{Schwarz, A.} \emph{et~al.}
\newblock \bibinfo{journal}{\bibinfo{title}{Characterization of stroke-related upper limb motor impairments across various upper limb activities by use of kinematic core set measures}}.
\newblock {\emph{\JournalTitle{Journal of neuroengineering and rehabilitation}}} \textbf{\bibinfo{volume}{19}}, \bibinfo{pages}{1--18} (\bibinfo{year}{2022}).

\bibitem{just2020human}
\bibinfo{author}{Just, F.} \emph{et~al.}
\newblock \bibinfo{journal}{\bibinfo{title}{Human arm weight compensation in rehabilitation robotics: efficacy of three distinct methods}}.
\newblock {\emph{\JournalTitle{Journal of neuroengineering and rehabilitation}}} \textbf{\bibinfo{volume}{17}}, \bibinfo{pages}{1--17} (\bibinfo{year}{2020}).

\bibitem{balasubramanian2015analysis}
\bibinfo{author}{Balasubramanian, S.}, \bibinfo{author}{Melendez-Calderon, A.}, \bibinfo{author}{Roby-Brami, A.} \& \bibinfo{author}{Burdet, E.}
\newblock \bibinfo{journal}{\bibinfo{title}{On the analysis of movement smoothness}}.
\newblock {\emph{\JournalTitle{Journal of neuroengineering and rehabilitation}}} \textbf{\bibinfo{volume}{12}}, \bibinfo{pages}{1--11} (\bibinfo{year}{2015}).

\bibitem{antonsson1985frequency}
\bibinfo{author}{Antonsson, E.~K.} \& \bibinfo{author}{Mann, R.~W.}
\newblock \bibinfo{journal}{\bibinfo{title}{The frequency content of gait}}.
\newblock {\emph{\JournalTitle{Journal of biomechanics}}} \textbf{\bibinfo{volume}{18}}, \bibinfo{pages}{39--47} (\bibinfo{year}{1985}).

\bibitem{radford2022whisper}
\bibinfo{author}{Radford, A.} \emph{et~al.}
\newblock \bibinfo{journal}{\bibinfo{title}{Robust speech recognition via large-scale weak supervision}}.
\newblock {\emph{\JournalTitle{arXiv preprint arXiv:2212.04356}}}  (\bibinfo{year}{2022}).

\bibitem{ffmpeg}
\bibinfo{author}{{FFmpeg Developers}}.
\newblock \bibinfo{title}{Ffmpeg}.
\newblock \bibinfo{howpublished}{\url{https://ffmpeg.org/}}.
\newblock \bibinfo{note}{Accessed: 2025-09-01}.

\bibitem{kiss2006unsupervised}
\bibinfo{author}{Kiss, T.} \& \bibinfo{author}{Strunk, J.}
\newblock \bibinfo{journal}{\bibinfo{title}{Unsupervised multilingual sentence boundary detection}}.
\newblock {\emph{\JournalTitle{Computational Linguistics}}} \textbf{\bibinfo{volume}{32}}, \bibinfo{pages}{485--525} (\bibinfo{year}{2006}).

\bibitem{reimers2019}
\bibinfo{author}{Reimers, N.} \& \bibinfo{author}{Gurevych, I.}
\newblock \bibinfo{title}{Sentence-bert: Sentence embeddings using siamese bert-networks}.
\newblock In \emph{\bibinfo{booktitle}{Proceedings of the 2019 Conference on Empirical Methods in Natural Language Processing}}, \bibinfo{pages}{3982--3992} (\bibinfo{year}{2019}).

\bibitem{mcauley1989psychometric}
\bibinfo{author}{McAuley, E.}, \bibinfo{author}{Duncan, T.} \& \bibinfo{author}{Tammen, V.~V.}
\newblock \bibinfo{journal}{\bibinfo{title}{Psychometric properties of the intrinsic motivation inventory in a competitive sport setting: A confirmatory factor analysis}}.
\newblock {\emph{\JournalTitle{Research quarterly for exercise and sport}}} \textbf{\bibinfo{volume}{60}}, \bibinfo{pages}{48--58} (\bibinfo{year}{1989}).

\bibitem{hart2006nasa}
\bibinfo{author}{Hart, S.~G.}
\newblock \bibinfo{title}{Nasa-task load index (nasa-tlx); 20 years later}.
\newblock In \emph{\bibinfo{booktitle}{Proceedings of the human factors and ergonomics society annual meeting}}, vol.~\bibinfo{volume}{50}, \bibinfo{pages}{904--908} (\bibinfo{organization}{Sage publications Sage CA: Los Angeles, CA}, \bibinfo{year}{2006}).

\bibitem{ratz2024designing}
\bibinfo{author}{R{\"a}tz, R.}, \bibinfo{author}{Ratschat, A.~L.}, \bibinfo{author}{Cividanes-Garcia, N.}, \bibinfo{author}{Ribbers, G.~M.} \& \bibinfo{author}{Marchal-Crespo, L.}
\newblock \bibinfo{journal}{\bibinfo{title}{Designing for usability: development and evaluation of a portable minimally-actuated haptic hand and forearm trainer for unsupervised stroke rehabilitation}}.
\newblock {\emph{\JournalTitle{Frontiers in neurorobotics}}} \textbf{\bibinfo{volume}{18}}, \bibinfo{pages}{1351700} (\bibinfo{year}{2024}).

\bibitem{kuznetsova2017lmertest}
\bibinfo{author}{Kuznetsova, A.}, \bibinfo{author}{Brockhoff, P.~B.} \& \bibinfo{author}{Christensen, R. H.~B.}
\newblock \bibinfo{journal}{\bibinfo{title}{lmertest package: tests in linear mixed effects models}}.
\newblock {\emph{\JournalTitle{Journal of statistical software}}} \textbf{\bibinfo{volume}{82}} (\bibinfo{year}{2017}).

\bibitem{Poggensee2021}
\bibinfo{author}{Poggensee, K.~L.} \& \bibinfo{author}{Collins, S.~H.}
\newblock \bibinfo{journal}{\bibinfo{title}{How adaptation, training, and customization contribute to benefits from exoskeleton assistance}}.
\newblock {\emph{\JournalTitle{Science Robotics}}} \textbf{\bibinfo{volume}{6}}, \bibinfo{pages}{eabf1078} (\bibinfo{year}{2021}).

\bibitem{sawers2014}
\bibinfo{author}{Sawers, A.} \& \bibinfo{author}{Ting, L.~H.}
\newblock \bibinfo{journal}{\bibinfo{title}{Perspectives on human–human sensorimotor interactions for the design of rehabilitation robots}}.
\newblock {\emph{\JournalTitle{Journal of NeuroEngineering and Rehabilitation}}} \textbf{\bibinfo{volume}{11}}, \bibinfo{pages}{142} (\bibinfo{year}{2014}).

\bibitem{DiTommaso2025}
\bibinfo{author}{Di~Tommaso, F.} \emph{et~al.}
\newblock \bibinfo{title}{Effects of visual and verbal feedback on active patient participation during robot-assisted gait training}.
\newblock In \emph{\bibinfo{booktitle}{2025 International Conference On Rehabilitation Robotics (ICORR)}}, \bibinfo{pages}{1281--1287} (\bibinfo{year}{2025}).

\bibitem{Michielsen14082019}
\bibinfo{author}{Michielsen, M.}, \bibinfo{author}{Vaughan-Graham, J.}, \bibinfo{author}{Holland, A.}, \bibinfo{author}{Magri, A.} \& \bibinfo{author}{Suzuki, M.}
\newblock \bibinfo{journal}{\bibinfo{title}{The bobath concept – a model to illustrate clinical practice}}.
\newblock {\emph{\JournalTitle{Disability and Rehabilitation}}} \textbf{\bibinfo{volume}{41}}, \bibinfo{pages}{2080--2092} (\bibinfo{year}{2019}).

\bibitem{Cucinella2024}
\bibinfo{author}{Cucinella, S.~L.} \emph{et~al.}
\newblock \bibinfo{title}{The value of active end-user participation in rehabilitation technology: A co-creation workshop}.
\newblock In \bibinfo{editor}{Pons, J.~L.}, \bibinfo{editor}{Tornero, J.} \& \bibinfo{editor}{Akay, M.} (eds.) \emph{\bibinfo{booktitle}{Converging Clinical and Engineering Research on Neurorehabilitation V}}, \bibinfo{pages}{636--640} (\bibinfo{publisher}{Springer Nature Switzerland}, \bibinfo{address}{Cham}, \bibinfo{year}{2024}).

\bibitem{Woodbridge2023}
\bibinfo{author}{Woodbridge, H.} \emph{et~al.}
\newblock \bibinfo{journal}{\bibinfo{title}{Clinician and patient perspectives on the barriers and facilitators to physical rehabilitation in intensive care: a qualitative interview study}}.
\newblock {\emph{\JournalTitle{BMJ Open}}} \textbf{\bibinfo{volume}{13}}, \bibinfo{pages}{e073061} (\bibinfo{year}{2023}).

\bibitem{Basalp2021}
\bibinfo{author}{Basalp, E.}, \bibinfo{author}{Wolf, P.} \& \bibinfo{author}{Marchal-Crespo, L.}
\newblock \bibinfo{journal}{\bibinfo{title}{Haptic training: Which types facilitate (re)learning of which motor task and for whom? answers by a review}}.
\newblock {\emph{\JournalTitle{IEEE Transactions on Haptics}}} \textbf{\bibinfo{volume}{14}}, \bibinfo{pages}{722--739} (\bibinfo{year}{2021}).

\bibitem{Brady2023}
\bibinfo{author}{Brady, N.}, \bibinfo{author}{Dejaco, B.}, \bibinfo{author}{Lewis, J.}, \bibinfo{author}{McCreesh, K.} \& \bibinfo{author}{McVeigh, J.~G.}
\newblock \bibinfo{journal}{\bibinfo{title}{Physiotherapist beliefs and perspectives on virtual reality supported rehabilitation for the management of musculoskeletal shoulder pain: A focus group study}}.
\newblock {\emph{\JournalTitle{PLOS ONE}}} \textbf{\bibinfo{volume}{18}}, \bibinfo{pages}{e0284445} (\bibinfo{year}{2023}).

\end{thebibliography}

\section*{Acknowledgements (not compulsory)}

The authors sincerely thank Nikki Korzilius for her contribution in the development of a proof-of-concept system for the haptic teleoperation system and Alberto Garz\'as-Villar for his support in the statistical analysis.

\section*{Author contributions statement}

M.L., A.R., A.vdB., and L.M.C. conceived the haptic framework and the experimental protocol, M.L. conducted the experiments, B.L. developed the speech-analysis framework, M.L. and B.L. analyzed the results, B.L. and L.M.C. drafted the manuscript. All authors reviewed the manuscript. 

\section*{Additional information}

 \textbf{Accession codes} data and codes will be available after publication on Zenodo; \\
 \textbf{Competing interests} All authors declare no competing interests. The corresponding author is responsible for submitting a \href{http://www.nature.com/srep/policies/index.html#competing}{competing interests statement} on behalf of all authors of the paper.

\end{document}


\section*{\Large Supplementary-Materials: Therapist-Robot-Patient Physical Interaction \textit{is Worth
a Thousand Words}: Enabling Intuitive Therapist
Guidance via Remote Haptic Control }
 Beatrice Luciani, Alex van den Berg, Matti Lang, Alexandre L. Ratschat, and Laura Marchal-Crespo

\section{Questionnaires}
\subsection{Participant Information Questionnaire} 

\begin{enumerate}
    \item \textbf{Participant Number:} \hrulefill

    \item \textbf{Name:} \hrulefill

    \item \textbf{Age:} \hrulefill

    \item \textbf{Sex assigned at birth:}
    \begin{itemize}
        \item Male
        \item Female
        \item Other (specify): \hrulefill
    \end{itemize}

    \item \textbf{Is English your mother tongue?}
    \begin{itemize}
        \item Yes
        \item No (specify): \hrulefill
    \end{itemize}

    \item \textbf{What is your English level?}
    \begin{itemize}
        \item A1 -- Beginner
        \item A2 -- Elementary
        \item B1 -- Pre-intermediate
        \item B2 -- Intermediate
        \item C1 -- Upper-intermediate
        \item C2 -- Advanced
        \item C3 -- Native speaker
    \end{itemize}

    \item \textbf{Handedness:}
    \begin{itemize}
        \item Right-handed
        \item Left-handed
        \item Ambidextrous
    \end{itemize}

    \item \textbf{Do you have previous experience with Immersive Virtual Reality?}
    \begin{itemize}
        \item Never used
        \item Used once or twice
        \item Used occasionally (a few times a year)
        \item Used frequently (several times a month)
        \item Used very frequently
    \end{itemize}

    \item \textbf{Do you have experience with the Sigma.7 device?}
    \begin{itemize}
        \item Never used
        \item Used once or twice
        \item Used occasionally (a few times a year)
        \item Used frequently (several times a month)
        \item Used very frequently
    \end{itemize}

    \item \textbf{Do you have experience with the ARMin Exoskeleton?}
    \begin{itemize}
        \item Never used
        \item Used once or twice
        \item Used occasionally (a few times a year)
        \item Used frequently (several times a month)
        \item Used very frequently
    \end{itemize}

    \item \textbf{Which role were you assigned for the experiment?}
    \begin{itemize}
        \item Leader
        \item Follower
    \end{itemize}
\end{enumerate}

\subsection{Final IMI and NASA questionnaire}
\subsection*{Intrinsic Motivation Inventory (IMI)}

\noindent\textbf{Role:} \quad Leader \quad $\square$ \qquad Follower \quad $\square$

\vspace{1em}
\setlength{\tabcolsep}{6pt}
\renewcommand{\arraystretch}{1.4}

\noindent\textbf{Perceived Competence}

\begin{center}
\begin{tabular}{p{7cm} c c c c c c c}
\hline
\textbf{Statement} &
\multicolumn{7}{c}{\textbf{Level of agreement}} \\
 & 1 & 2 & 3 & 4 & 5 & 6 & 7 \\
 & \scriptsize Not at all true & & & \scriptsize Somewhat true & & & \scriptsize Very true \\
\hline
I think I am pretty good at this activity
& $\circ$ & $\circ$ & $\circ$ & $\circ$ & $\circ$ & $\circ$ & $\circ$ \\

I think I did pretty well at this activity, compared to other students
& $\circ$ & $\circ$ & $\circ$ & $\circ$ & $\circ$ & $\circ$ & $\circ$ \\

After working at this activity for a while, I felt pretty competent
& $\circ$ & $\circ$ & $\circ$ & $\circ$ & $\circ$ & $\circ$ & $\circ$ \\

I am satisfied with my performance at this task
& $\circ$ & $\circ$ & $\circ$ & $\circ$ & $\circ$ & $\circ$ & $\circ$ \\

I was pretty skilled at this activity
& $\circ$ & $\circ$ & $\circ$ & $\circ$ & $\circ$ & $\circ$ & $\circ$ \\

This was an activity that I couldn't do very well
& $\circ$ & $\circ$ & $\circ$ & $\circ$ & $\circ$ & $\circ$ & $\circ$ \\
\hline
\end{tabular}
\end{center}

\vspace{1em}
\noindent\textbf{Effort}

\begin{center}
\begin{tabular}{p{7cm} c c c c c c c}
\hline
\textbf{Statement} &
\multicolumn{7}{c}{\textbf{Level of agreement}} \\
 & 1 & 2 & 3 & 4 & 5 & 6 & 7 \\
 & \scriptsize Not at all true & & & \scriptsize Somewhat true & & & \scriptsize Very true \\
\hline
I put a lot of effort into this
& $\circ$ & $\circ$ & $\circ$ & $\circ$ & $\circ$ & $\circ$ & $\circ$ \\

I didn't try very hard to do well at this activity
& $\circ$ & $\circ$ & $\circ$ & $\circ$ & $\circ$ & $\circ$ & $\circ$ \\

I tried very hard on this activity
& $\circ$ & $\circ$ & $\circ$ & $\circ$ & $\circ$ & $\circ$ & $\circ$ \\

It was important to me to do well at this task
& $\circ$ & $\circ$ & $\circ$ & $\circ$ & $\circ$ & $\circ$ & $\circ$ \\

I didn't put much energy into this
& $\circ$ & $\circ$ & $\circ$ & $\circ$ & $\circ$ & $\circ$ & $\circ$ \\
\hline
\end{tabular}
\end{center}

\subsection*{NASA Task Load Index (NASA-TLX)}

\noindent\textbf{Instructions:} Please indicate your rating for each dimension on a scale from 0 (Very Low) to 20 (Very High) by marking the bar.

\vspace{0.5em}
\setlength{\tabcolsep}{3pt}
\renewcommand{\arraystretch}{1.6}

\begin{center}
\begin{tabular}{p{5cm} l}
\hline
\textbf{Dimension / Question} & \textbf{Rating Bar (0--20)} \\
\hline
Mental Demand: How mentally demanding was the task? & 
0 \underline{\hspace{12cm}} 20 \\

Physical Demand: How physically demanding was the task? & 
0 \underline{\hspace{12cm}} 20 \\

Temporal Demand: How hurried or rushed was the pace of the task? & 
0 \underline{\hspace{12cm}} 20 \\

Performance: How successful were you in accomplishing what you were asked to do? & 
0 \underline{\hspace{12cm}} 20 \\

Effort: How hard did you have to work to accomplish your level of performance? & 
0 \underline{\hspace{12cm}} 20 \\

Frustration: How insecure, discouraged, irritated, stressed, and annoyed were you? & 
0 \underline{\hspace{12cm}} 20 \\
\hline
\end{tabular}
\end{center}

\section{Interview questions}
Questions for the Trainer:
\begin{itemize}
    \item Did you follow any strategy when using the Haptic device?
    \item Was the virtual environment useful or confusing?
\end{itemize}
Questions for the Trainee:
\begin{itemize}
    \item Did you feel safe?
    \item Was the experiment physically challenging?
\end{itemize}
Questions for both the Trainer and the Trainees:
\begin{itemize}
    \item Which condition did you prefer?
    \item Did you feel in control of the execution of the movements?
    \item Did you have any communication problems?
    \item Do you have any extra comments or remarks?
\end{itemize}

\section{Statistical analysis results}
Outliers were identified and removed using the following criterion: values below $Q1 - 2\cdot \text{IQR}$ or above $Q3 + 2 \cdot \text{IQR}$, where $Q1$ and $Q3$ are the first and third quartiles, respectively. Across all dyads and conditions, 1.25-5.42\% of observations were excluded per metric (see Tab.~\ref{tab:outliers}). Removal reduced the mean values by 6-14\% and decreased the standard deviation, reflecting the effect of extreme values. The Wilcoxon rank-sum test indicated that the distribution of the remaining data was not significantly different from the original ($p > 0.05$), confirming that the central tendencies of the data were preserved.
\begin{table}[H]
\centering
\caption{Outlier removal per metric and demonstration type. Percentages indicate the proportion of observations excluded using the $2.0 \cdot$ IQR criterion. Means and standard deviations are reported before and after outlier removal.}
\label{tab:outliers}
\begin{tabular}{lllllllll}
\hline
Metric & Condition & Original & Removed Out. & \% Out. & Mean Before & Mean After & SD Before & SD After \\
\hline
\multirow{2}{*}{Completion Time} 
 & VD        & 240 & 8 &  3.33 & 22.6 & 20.4 & 16.5 & 9.94 \\
 & HD        & 240 & 13 & 5.42 & 21.3 & 18.3 & 16.3 & 8.83 \\
\hline
\multirow{4}{*}{Smoothness} 
 & VD (Elbow)  & 240 & 7 & 2.92 & -4.20 & -3.92 &  2.44   &  1.83\\
  & VD (Hand)  & 240 & 9 & 3.75 & -3.95 & -3.67   & 2.15   &   1.58 \\
 & HD (Elbow) & 240 & 3 & 1.25 & -3.87 &  -3.76  & 2.15   & 1.90 \\
 & HD (Hand) & 240 & 4 & 1.67 & -3.74 & -3.62 & 2.02   &  1.82\\
\hline
\end{tabular}
\end{table}

Before selecting the final model for our analyses, we compared alternative formulations. In the final model, the effect of the different poses on the quantitative outcomes was not included, as it was not relevant for addressing the main research questions of the study. For completeness, however, we report here the models that were built including the factor \textbf{Pose}. When comparing these extended models with the final models presented in the paper, we found that removing \textbf{Pose} did not lead to a substantial increase in either the BIC or AIC indices.
\subsection{Completion Time}
\begin{figure}[h!]
         \centering
    \includegraphics[width=\textwidth]{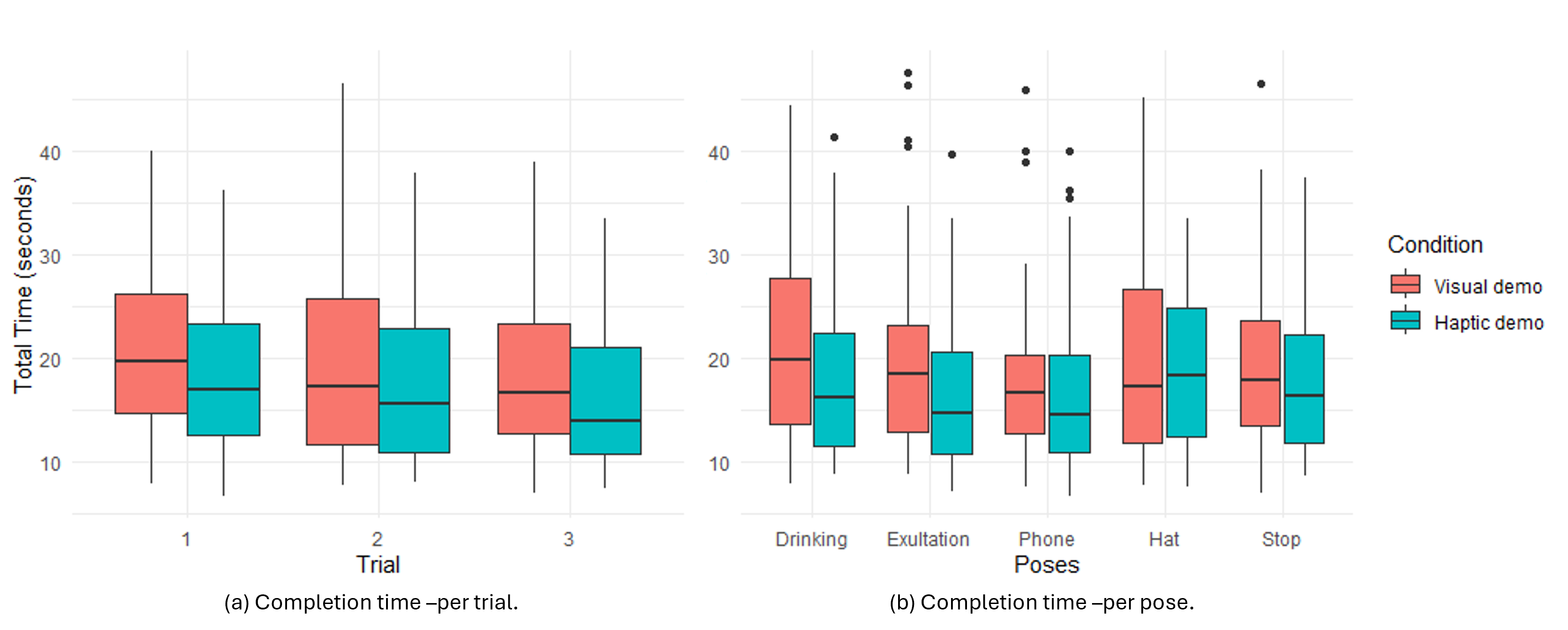}
     \caption{(a) Statistics of the time needed to complete the poses in the two conditions, along with the trials. (b) Statistics of the time needed to complete the poses in the two conditions, divided by pose.}
     \label{fig: time_results}
\end{figure}

\begin{table}[H] 
\centering 
\resizebox{0.5\textwidth}{!}{ 
\begin{tabular}{lllllll} 
\toprule 
& Estimate & Std. Error & df & t value & Pr($>|t|$) & \\ 
\midrule
(Intercept) & 19.21 & 1.92 & 39.16 & 9.98 & 2.56e-12 & *** \\
Haptic & -3.43 & 1.89 & 421.23 & -1.82 & 0.0698 & . \\ 
Pose\_Exult & 2.15 & 1.53 & 421.13 & 1.40 & 0.1621 & \\ 
Pose\_Phone & 3.33 & 1.54 & 421.17 & 2.16 & 0.0312 & * \\
Pose\_Stop & 1.49 & 1.54 & 421.24 & 0.97 & 0.3350 & \\ 
Pose\_Hat & 1.36 & 1.52 & 421.11 & 0.89 & 0.3716 & \\ 
Order2 & -0.16 & 2.21 & 17.22 & -0.07 & 0.9430 & \\ 
Trial & -0.89 & 0.60 & 421.28 & -1.48 & 0.1399 & \\ 
Haptic:Pose\_Exult & -1.60 & 2.19 & 421.25 & -0.73 & 0.4642 & \\ 
Haptic:Pose\_Phone & -1.19 & 2.20 & 421.20 & -0.54 & 0.5879 & \\ 
Haptic:Pose\_Stop & 0.03 & 2.19 & 421.10 & 0.01 & 0.9902 & \\ 
Haptic:Pose\_Hat & 0.57 & 2.16 & 421.13 & 0.26 & 0.7933 & \\ 
Haptic:Order2 & 4.58 & 1.40 & 421.20 & 3.28 & 0.00112 & ** \\ 
Haptic:Trial & -0.40 & 0.86 & 421.56 & -0.46 & 0.6432 & \\ 
\bottomrule 
\end{tabular} 
} 
\caption{Results of the Completion Time LMM. *($p <0.05$), **($p <0.01$), ***($p <0.001$)} 
\label{tab:time} 
\end{table}

\subsection{Smoothness}
\begin{table}[H]
\centering
\resizebox{0.95\textwidth}{!}{
\begin{tabular}{lcccccc|cccccc}
\toprule
& \multicolumn{6}{c}{\textbf{Elbow smoothness [-]}} & \multicolumn{6}{c}{\textbf{Wrist smoothness [-]}} \\ 
\cmidrule(lr){2-7} \cmidrule(lr){8-13}
 & Estimate & Std. Error & df & t value & Pr($>|t|$) &  & Estimate & Std. Error & df & t value & Pr($>|t|$) & \\ 
\midrule
(Intercept)       & -4.18 & 0.68 & 32.96 & -6.17 & 5.83e-07 & *** & -3.98 & 0.35 & 51.98 & -11.54 & 5.95e-16 & *** \\
Haptic            & 1.23 & 0.64 & 461.39 & 1.92 & 0.056 & . & 0.13 & 0.35 & 460.31 & 0.36 & 0.723 &  \\
Pose\_Exult       & 0.13 & 0.31 & 461.23 & 0.43 & 0.668 &  & 0.01 & 0.29 & 460.38 & 0.038 & 0.970 &  \\
Pose\_Phone       & 0.06 & 0.31 & 461.23 & 0.20 & 0.839 &  & 0.48 & 0.29 & 460.39 & 1.652 & 0.099 & . \\
Pose\_Hat         & 0.20 & 0.31 & 461.23 & 0.65 & 0.515 &  & 0.34 & 0.28 & 460.26 & 1.188 & 0.235 &  \\
Pose\_Stop        & 0.02 & 0.31 & 461.05 & 0.054 & 0.957 &  & -0.046 & 0.285 & 460.17 & -0.160 & 0.873 &  \\
Order2            & -0.10 & 0.37 & 20.28 & -0.278 & 0.784 &  & 0.030 & 0.368 & 19.43 & 0.082 & 0.935 &  \\
Trial             & 0.201 & 0.121 & 461.20 & 1.659 & 0.098 & . & 0.199 & 0.111 & 460.26 & 1.793 & 0.074 & . \\
Haptic:Pose\_Exult & 0.095 & 0.448 & 461.42 & 0.211 & 0.833 &  & 0.566 & 0.405 & 460.28 & 1.399 & 0.163 &  \\
Haptic:Pose\_Phone & 0.009 & 0.443 & 461.17 & 0.019 & 0.985 &  & 0.230 & 0.404 & 460.19 & 0.568 & 0.570 &  \\
Haptic:Pose\_Hat   & -0.004 & 0.443 & 461.16 & -0.008 & 0.993 &  & 0.505 & 0.400 & 460.19 & 1.262 & 0.208 &  \\
Haptic:Pose\_Stop  & 0.119 & 0.442 & 461.09 & 0.269 & 0.788 &  & 0.337 & 0.399 & 460.13 & 0.843 & 0.400 &  \\
Haptic:Order2      & -0.640 & 0.281 & 461.24 & -2.280 & 0.023 & * & -0.711 & 0.254 & 460.22 & -2.801 & 0.005 & ** \\
Haptic:Trial       & -0.005 & 0.173 & 461.20 & -0.027 & 0.979 &  & -0.015 & 0.156 & 460.18 & -0.098 & 0.922 &  \\
\bottomrule
\end{tabular}
}
\caption{Results of the LMMs for the smoothness at elbow and wrist levels.  
*($p < 0.05$), **($p < 0.01$), ***($p < 0.001$), .($p < 0.1$)}
\label{tab: smooth}
\end{table}

\section{Questionnaires results}

According to the questionnaire results for the NASA-TLX, Trainers experienced a significantly higher mental demand than Trainees. The HD condition, however, did not significantly alter the mental demand of the global population compared to the VD condition.

In contrast, if being a Trainer did not significantly increase the perceived physical demand with respect to being a Trainee, Trainees globally perceived less physical demand during the HD condition.
Lastly, our participants reported significantly lower perceived effort in the HD condition, with a non-significant effect for the role × condition interaction.
Analyzing the temporal demand, performance, and frustration subcategories, we did not detect any significant effect due to the Role or the Condition.
\begin{table*}[h!]
\centering
\scriptsize
\begin{minipage}[t]{0.48\textwidth}
\centering
\resizebox{\linewidth}{!}{
\begin{tabular}{llccccccc}
\toprule
\textbf{Outcome} & \textbf{Term} & \textbf{Estimate} & \textbf{Std. Error} & \textbf{95\% CI} & \textbf{df} & \textbf{Pr($>|t|$)} & \textbf{$p_{Holm}$} \\
\midrule
\multirow{4}{*}{\makecell{\textbf{Mental} \\ \textbf{Demand}}} 
& (Intercept) & 22.50 & 5.27 & [12.09, 32.91] & 51.36 & 8.39e-05 *** & 3.36e-04 *** \\
& Trainer & 30.00 & 7.45 & [15.82, 44.18] & 51.36 & 1.86e-04 *** & 3.73e-04 *** \\
& HD & -0.94 & 5.72 & [-12.14, 10.26] & 30.00 & 0.871 & 0.871\\
& Trainer:HD & -3.44 & 8.09 & [-19.45, 12.57] & 30.00 & 0.674 & 0.871\\ \midrule
\multirow{4}{*}{\makecell{\textbf{Physical} \\ \textbf{Demand}}} 
& (Intercept) & 55.31 & 6.04 & [43.47, 67.15] & 59.25 & 6.00e-13 *** & 2.40e-12*** \\
& Trainer & -7.81 & 8.54 & [-24.65, 9.03] & 59.25 & 0.364&0.364\\
& HD & -20.31 & 8.05 & [-36.05, -4.57] & 30.00 & 0.017 * & 0.034* \\
& Trainer:HD & 14.69 & 11.38 & [-7.70, 37.08] & 30.00 & 0.207& 0.275\\ \midrule
\multirow{4}{*}{\makecell{\textbf{Temporal} \\ \textbf{Demand}}} 
& (Intercept) & 20.63 & 5.05 & [10.76, 30.50] & 43.19 & 1.86e-04*** &7.45e-04*** \\
& Trainer & 10.94 & 7.14 & [-3.08, 24.96] & 43.19 & 0.133 &0.265\\
& HD & 4.06 & 4.38 & [-4.53, 12.65] & 30.00 & 0.361&0.481 \\
& Trainer:HD & -2.50 & 6.19 & [-14.67, 9.67] & 30.00 & 0.689 &0.689\\ \midrule
\multirow{4}{*}{\textbf{Performance}} 
& (Intercept) & 76.25 & 4.20 & [68.01, 84.49] & 55.98 & <2e-16*** &<2e-16*** \\
& Trainer & -1.56 & 5.94 & [-13.28, 10.16] & 55.98 & 0.793 & 0.793\\
& HD & -2.81 & 5.08 & [-12.83, 7.21] & 30.00 & 0.584& 0.779 \\
& Trainer:HD & 5.94 & 7.19 & [-8.38, 20.26] & 30.00 & 0.415 &0.779\\ \midrule
\multirow{4}{*}{\textbf{Effort}} 
& (Intercept) & 53.13 & 5.78 & [41.80, 64.46] & 58.76 & 5.86e-13*** &2.34e-12*** \\
& Trainer & 7.50 & 8.18 & [-8.62, 23.62] & 58.76 & 0.363 &0.363\\
& HD & -28.44 & 7.56 & [-43.28, -13.60] & 30.00 & 7.36e-04*** &0.001** \\
& Trainer:HD & 18.75 & 10.70 & [-2.35, 39.85] & 30.00 & 0.090.& 0.120 \\ \midrule
\multirow{4}{*}{\textbf{Frustration}} 
& (Intercept) & 10.31 & 5.17 & [0.11, 20.51] & 60.00 & 0.0506.& 0.201 \\
& Trainer & 12.19 & 7.31 & [-2.33, 26.71] & 60.00 & 0.101 &0.201\\
& HD & 1.88 & 7.31 & [-12.52, 16.28] & 60.00 & 0.798 &0.798\\
& Trainer:HD & -4.06 & 10.34 & [-24.33, 16.21] & 60.00 & 0.696 &0.798\\
\bottomrule
\multicolumn{7}{l}{\scriptsize Significance codes: *** $p < .001$, ** $p < .01$, * $p < .05$, . $p < .1$}
\end{tabular}
}
\caption{Linear Mixed Model results for NASA-TLX subscales.}
\label{Tab:NASA}
\end{minipage}
\hfill
\begin{minipage}[t]{0.48\textwidth}
\centering
\resizebox{\linewidth}{!}{
\begin{tabular}{llccccccc}
\toprule
\textbf{Outcome} & \textbf{Term} & \textbf{Estimate} & \textbf{Std. Error} & \textbf{95\% CI} & \textbf{df} & \textbf{Pr($>|t|$)} & \textbf{$p_{Holm}$}\\
\midrule
\multirow{4}{*}{\makecell{\textbf{Perceived} \\ \textbf{Competence}}} 
& (Intercept) & 70.83 & 4.01 & [62.99, 78.67] & 38.67 & $< 2\times10^{-16}$ *** & $< 2\times10^{-16}$ ***\\
& Trainer & -4.34 & 5.67 & [-15.45, 6.77] & 38.67 & 0.449 & 0.598\\
& HD & -0.69 & 2.42 & [-5.44, 4.06] & 352.00 & 0.774&  0.774\\
& Trainer:HD & 10.59 & 3.42 & [3.89, 17.29] & 352.00 & 0.002 **& 0.004 ** \\
\midrule
\multirow{4}{*}{\textbf{Effort}} 
& (Intercept) & 73.33 & 4.64 & [64.22, 82.44] & 42.79 & $< 2\times10^{-16}$ ***  & $< 2\times10^{-16}$ *** \\
& Trainer & 5.00 & 6.56 & [-8.87, 18.87] & 42.79 & 0.450 &0.450\\
& HD & -15.42 & 3.43 & [-21.97, -8.87] & 288.00 & 9.98e-06 ***& 1.99e-05*** \\
& Trainer:HD & 11.88 & 4.85 & [2.37, 21.39] & 288.00 & 0.015 * & 0.020 * \\
\bottomrule
\multicolumn{7}{l}{\scriptsize Significance codes: *** $p < .001$, ** $p < .01$, * $p < .05$, . $p < .1$}
\end{tabular}
}
\caption{Linear Mixed Model results for IMI questionnaire subscales.}
\label{Tab:IMI}
\end{minipage}
\end{table*}